\documentclass[10pt,twocolumn,letterpaper]{article}

\usepackage[pagenumbers]{cvpr} 

\usepackage{graphicx}
\usepackage{amsmath}
\usepackage{amssymb}
\usepackage{booktabs}
\usepackage{pifont}
\usepackage{xspace}
\usepackage{multirow}
\usepackage{colortbl}

\usepackage[pagebackref=true,breaklinks=true,letterpaper=true,colorlinks,citecolor=blue,linkcolor=red,bookmarks=false]{hyperref}

\usepackage[capitalize]{cleveref}
\crefname{section}{Sec.}{Secs.}
\Crefname{section}{Section}{Sections}
\Crefname{table}{Table}{Tables}
\crefname{table}{Tab.}{Tabs.}

\def\cvprPaperID{470} 
\def\confName{CVPR}
\def\confYear{2023}

\begin{document}
	\title{Spatially-Adaptive Feature Modulation for Efficient Image Super-Resolution}
	\author{~~~~Long Sun, Jiangxin Dong, Jinhui Tang, Jinshan Pan~~~~~\\
		~~~~Nanjing University of Science and Technology~~~~\\}

\maketitle

\begin{abstract}
	Although numerous solutions have been proposed for image super-resolution, they are usually incompatible with low-power devices with many computational and memory constraints.
	In this paper, we address this problem by proposing a simple yet effective deep network to solve image super-resolution efficiently. 
	In detail, we develop a spatially-adaptive feature modulation (SAFM) mechanism upon a vision transformer (ViT)-like block.
	Within it, we first apply the SAFM block over input features to dynamically select representative feature representations.
	As the SAFM block processes the input features from a long-range perspective, we further introduce a convolutional channel mixer (CCM) to simultaneously extract local contextual information and perform channel mixing.
	Extensive experimental results show that the proposed method is \textbf{3}$\times$ smaller than state-of-the-art efficient SR methods, e.g., IMDN, in terms of the network parameters and requires less computational cost while achieving comparable performance.
    The code is available at \url{https://github.com/sunny2109/SAFMN}.
\end{abstract}
\vspace{-5mm}
\section{Introduction}
\label{sec:intro}
Single image super-resolution (SISR) aims to restore a high-resolution (HR) image from its low-resolution (LR) counterpart by recovering lost details. 
This longstanding and challenging task has recently attracted much attention due to the rapid development of streaming media or high-definition devices. 
As these scenarios are usually resource-constrained,
it is of great interest to develop an efficient and effective SR method to estimate HR images for better visual display on these platforms or products.

\begin{figure}
	\centering
	\includegraphics[width=0.49\textwidth]{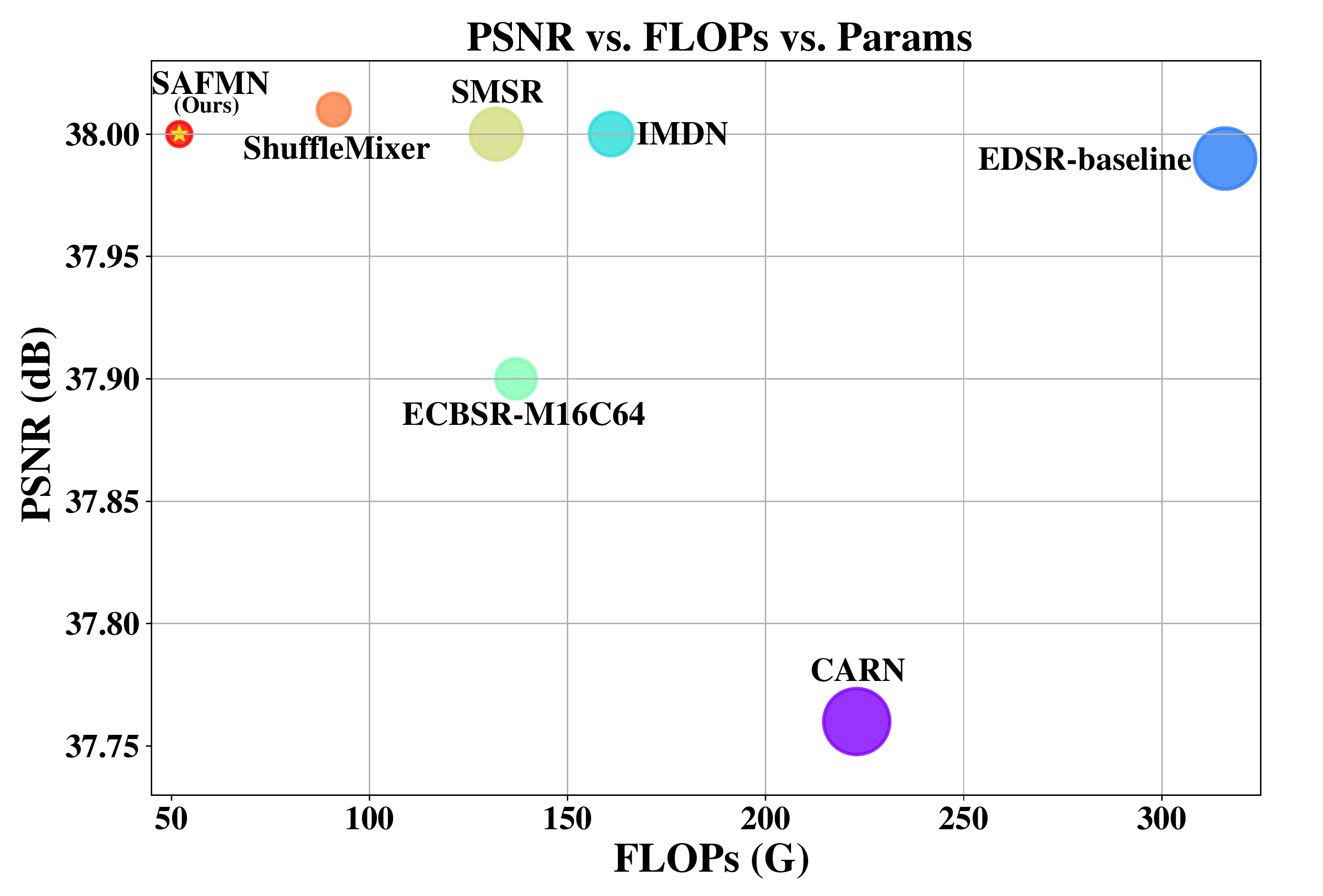}
	\vspace{-6mm}
	\caption{Model complexity and performance comparison between our proposed SAFMN model and other lightweight methods on Set5~\cite{Set5} for $\times2$ SR. Circle sizes indicate the number of parameters. The proposed method achieves a better trade-off between model complexity and reconstruction performance.}  
	\label{fig:comp}
	\vspace{-7mm}
\end{figure}

Deep learning-based SR methods have achieved significant performance improvements with the great evolution of hardware technologies, 
as we can use large amounts of data to train much larger or deeper neural networks for image SR~\cite{EDSR, RCAN, SwinIR, IPT}.
For example, RCAN~\cite{RCAN} is a representative CNN-based image SR network with 15.59M parameters and reaching a depth of over 400 layers.
One of the most significant drawbacks of these large models is that they require high computational costs, which makes them challenging to deploy. 
Moreover, recent visual transformers (ViTs)~\cite{vit, SwinIR, IPT} well beyond convolutional neural networks (CNNs) in low-level vision tasks, and their results demonstrate that exploring non-local feature interactions is essential for high-quality reconstruction.
But existing self-attention mechanisms are computationally expensive and unfriendly to efficient SR design.
This, therefore, motivates us to develop a lightweight yet effective model for real-world applications of image super-resolution by integrating the principles of convolution and self-attention.

To reduce the heavy computational burden, various methods, including efficient module design~\cite{FSRCNN, eSR, CARN, IMDN, rlfn, rfdn, ShuffleMixer, PAN, BSRN}, knowledge distillation~\cite{FAKD}, neural architecture search~\cite{FALSR, NASSR}, and structural re-parameterization~\cite{ECBSR}, are trying to improve the efficiency of SR algorithms. 
Among these efficient SR models, one direction is to reduce model parameters or complexity (FLOPs). 
Lightweight strategies like recursive manner~\cite{DRCN, DRRN}, parameter sharing~\cite{CARN}, and spare convolutions~\cite{CARN,ShuffleMixer,BSRN} are adopted. 
Although these approaches certainly reduce the model size, 
they usually compensate for the performance drop caused by shared recursive modules or sparse convolutions by increasing the depth or width of the model, 
which affects the inference efficiency when performing SR reconstruction.

Another direction is to accelerate the inference time.
The post-upsampling~\cite{FSRCNN, ESPCN} is an important replacement for the pre-defined input~\cite{srcnn,VDSR}, which significantly speeds up the runtime.
Model quantization~\cite{mobileai2022} effectively accelerates latency and reduces energy consumption, particularly when deploying algorithms in edge-devices.
Structural re-parameterization~\cite{ECBSR,repvgg} improves the speed of a well-trained model in the inference stage.
These methods enjoy fast running time but poor reconstruction performance.
Consequently, there is still room for a better trade-off between model efficiency and reconstruction performance.

\begin{figure}
    \centering
    \includegraphics[width=0.18\linewidth]{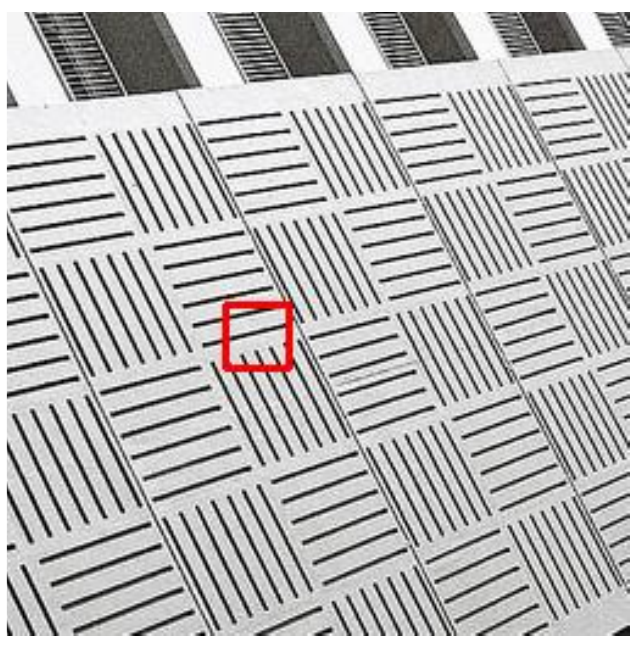}
    \includegraphics[width=0.18\linewidth]{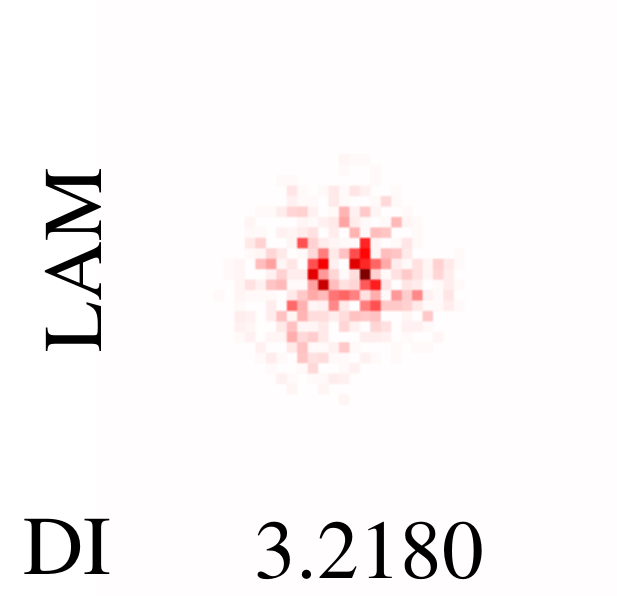}
    \includegraphics[width=0.18\linewidth]{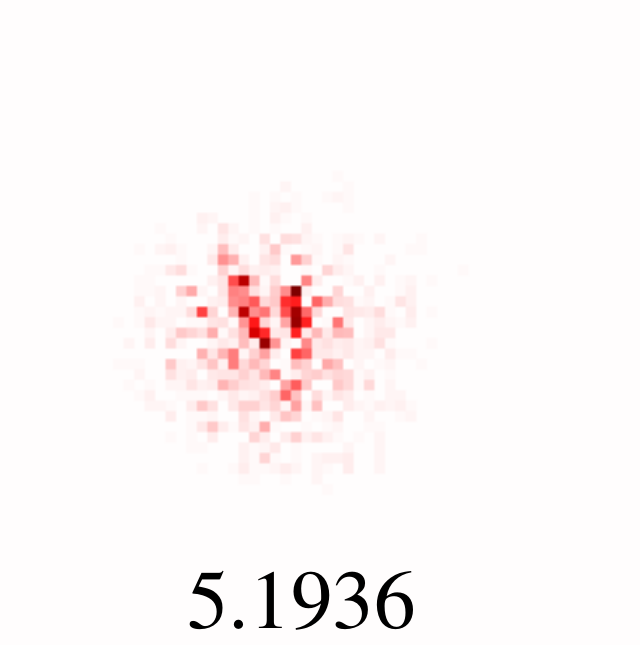}
    \includegraphics[width=0.18\linewidth]{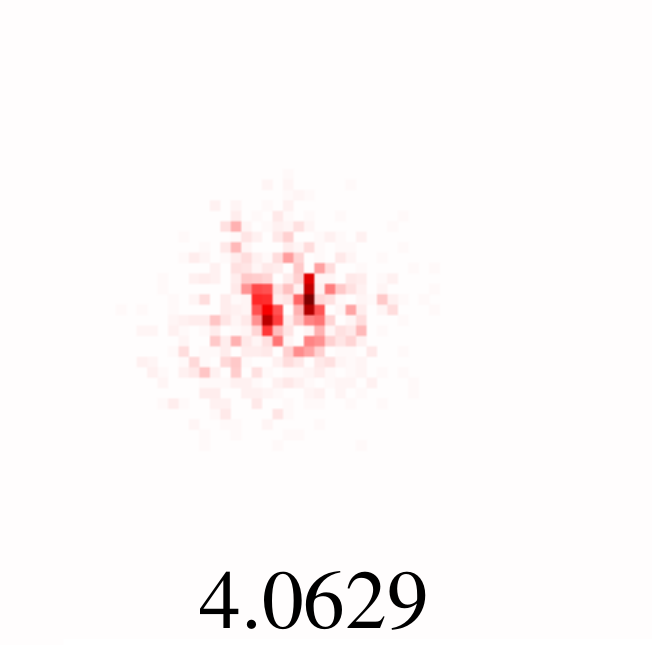}
    \includegraphics[width=0.18\linewidth]{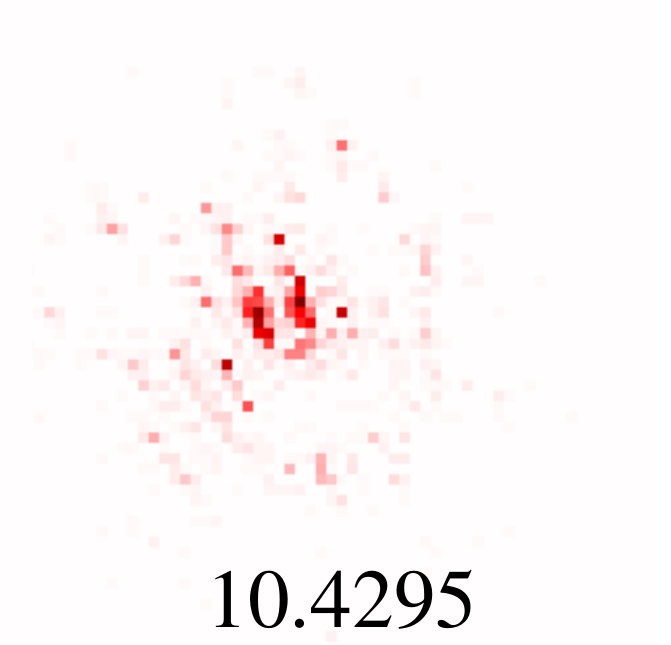}
    \\
    \makebox[0.185\linewidth]{\tiny (a) LR input}
    \makebox[0.185\linewidth]{\tiny (b) CARN~\cite{CARN} }
    \makebox[0.185\linewidth]{\tiny (c) EDSR-baseline~\cite{EDSR}}
    \makebox[0.185\linewidth]{\tiny (d) PAN~\cite{PAN}}
    \makebox[0.185\linewidth]{\tiny (e) SAFMN}
    \\
    \caption{Comparison of local attribution maps (LAMs) ~\cite{LAM} and diffusion indices (DIs) ~\cite{LAM} between our SAFMN and other efficient SR models.
    The LAM results denote the importance of each pixel in the input LR image when super-resolving the patch marked with a red box.
    The DI value reflects the range of involved pixels. 
    A larger DI value means a wider range of attention.
    The proposed method could exploit more feature information.
    }
    \label{fig:LAM}
    \vspace{-5mm}
\end{figure}

To address the above-mentioned issues, we design a simple yet effective model by developing a spatially-adaptive feature modulation, namely SAFMN, to realize a favorable trade-off between performance and efficiency.
Different from stacking lightweight convolutional modules, we explore a ViT-like architecture for better modeling long-range feature relations, as depicted in Figure~\ref{fig:LAM}.
Specifically, we develop a multi-scale representation-based feature modulation mechanism to dynamically select representative features.
Since the modulation mechanism processes input features from a long-range perspective, 
there is a requirement to complement local contextual information.
To this end, we present a convolutional channel mixer based on FMBConv~\cite{EfficientNetv2} to encode local features and mix channels simultaneously.
Taken together, we find that the SAFMN network is able to achieve a better trade-off between SR performance and model complexity, as shown in Figure~\ref{fig:comp}.

The main contributions of this paper are summarized as follows:
\begin{itemize}
	\vspace{-2mm}
	\item We propose a lightweight and effective SR model which
    absorbs CNN-like efficiency and transformer-like adaptability.
	\vspace{-2mm}
	\item We develop a compact convolutional channel mixer to encode local contextual information and perform channel mixing simultaneously.
	\vspace{-2mm}
	\item We evaluate the proposed method quantitatively and qualitatively on benchmark datasets, and the results show that our SAFMN achieves a favorable trade-off between accuracy and model complexity.
\end{itemize}

\section{Related Work}
\label{sec:related_work}
\noindent
\textbf{Deep Learning-based Image Super-Resolution.}
Classical interpolation algorithms, such as linear or bicubic upsampling, create high-resolution images by inserting zeros between adjacent pixels in the low resolution and then using a low-pass filter to preserve the content information of the input image~\cite{eSR}.
Unlike these interpolation-based upsamplers, deep learning-based approaches learn a nonlinear mapping between the input image and the target output in an end-to-end training fashion.
SRCNN~\cite{srcnn} is the first attempt to use a convolutional neural network (CNN) to tackle the image SR problem and achieves a considerable performance gain compared with conventional methods. 
Since then, many improvements have been proposed.
VDSR~\cite{VDSR} uses global residual learning~\cite{ResNet} to solve the problem of difficulty in training an SR model with deep layers.
DRRN~\cite{DRRN} integrates the local residual learning and global residual connection to ease the training difficulty and enhance high-frequency details.
EDSR~\cite{EDSR} further increases the model footprint to 43M, achieving a significant breakthrough in reconstruction performance and showing that the BatchNorm (BN)~\cite {BN} layer is not necessary for the SR task.
RCAN~\cite{RCAN} builds a more than 400 layers model based on channel attention and dense connections for accurate SR.
With the successful application of ViT in various high-level vision tasks, image SR also follows this ViT~\cite{vit} scheme and obtains higher performance than CNN-based models.
For instance, SwinIR~\cite{SwinIR} acts as a strong baseline for image restoration tasks based on the Swin Transformer~\cite{Swin}.
While these approaches achieve impressive reconstruction performance, the required high computational costs make them challenging to deploy in real-world applications on resource-constrained devices.

\begin{figure*}[!t]
	\footnotesize
	\centering
	\begin{tabular}{c}
		\includegraphics[width=\textwidth]{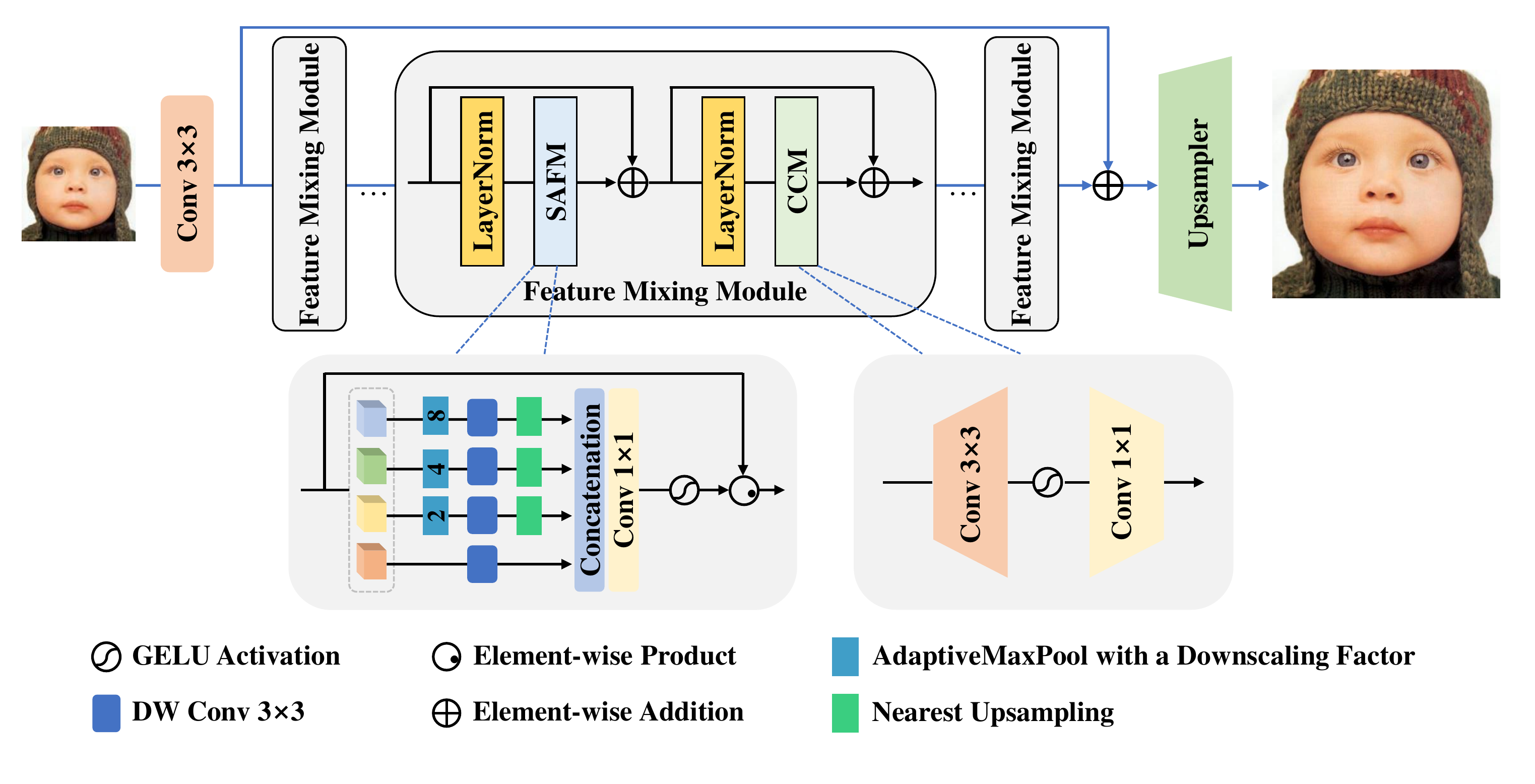}\\
	\end{tabular}
	\vspace{-3mm}
	\caption{An overview of the proposed SAFMN. SAFMN first transforms the input LR image into the feature space using a convolutional layer, performs feature extraction using a series of feature mixing modules (FMMs), and then reconstructs these extracted features by an upsampler module. The FMM block is mainly implemented by a spatially-adaptive feature modulation (SAFM) layer and a convolutional channel mixer (CCM).}  
	\label{fig:arc}
	\vspace{-5mm}
\end{figure*}

\noindent
\textbf{Efficient Image Super-Resolution.}
To improve the model efficiency, many CNN-based SR works try to alleviate this issue.
FSRCNN~\cite{FSRCNN} and ESPCN~\cite{ESPCN} utilize a post-upsampling manner to reduce the computational burden from the pre-defined inputs significantly.
CARN~\cite{CARN} uses group convolutions and a cascading mechanism upon residual networks to improve efficiency. 
IMDN~\cite{IMDN} adopts feature splitting and concatenation operations to progressively aggregate features, and its improved variants~\cite{rfdn, rlfn} won the AIM2020 and NTIRE2022 Efficient SR challenge. 
ShuffleMixer~\cite{ShuffleMixer} introduces a large kernel convolution for lightweight SR design.
BSRN~\cite{BSRN} proposes a blueprint-separable convolution-based model to reduce model complexity.
Meanwhile, an increasingly popular direction is to compress or accelerate a well-trained deep model through model quantization~\cite{mobileai2022}, structural re-parameterization~\cite{ECBSR} or knowledge distillation~\cite{FAKD}.
Neural architecture search (NAS) is also commonly used to search a well-constrained architecture for image super-resolution~\cite{NASSR, FALSR}. 
Note that the efficiency of a deep neural network could be measured in different metrics, including the number of parameters, FLOPs, activations, memory consumption and inference running time~\cite{ntire22efficientsr, AIM19constrainedSR}.
Although the above approaches have been improved in different efficiency aspects, there is still room for a favorable trade-off between reconstruction performance and model efficiency.
\vspace{-1mm}

\section{Proposed Method}
\label{sec:method}
In this section, we present the core components of our proposed model for efficient SISR.
As shown in Figure~\ref{fig:arc}, the network consists of the following parts: a stacking of feature mixing modules (FMMs) and an upsampler layer.
Specifically, we first apply a convolution layer with a kernel size of $3 \times 3$ pixels to transform the input LR image to feature space and generate the shallow feature $\text{F}_0$.
Then, the multiple stacked FMMs are used to generate finer deep features from $\text{F}_0$ for HR image reconstruction, where an FMM layer has a spatially-adaptive feature modulation (SAFM) sub-layer and a convolutional channel mixer (CCM).
To recover the HR target image, we introduce a global residual connection to learn high-frequency details and employ a lightweight upsampling layer for fast reconstruction, which only contains a $3 \times $3 convolution and a sub-pixel convolution~\cite{ESPCN}.
Our network can be formulated as:
\begin{equation}
	\begin{split}
		& \text{F}_0 = \mathcal{C}_\omega(I_{LR}), \\
		& I_{SR} = \mathcal{U}_\gamma(\mathcal{M}_\theta(\text{F}_0) + \text{F}_0),
	\end{split}
	\label{eqa:arc}
\end{equation}
where $I_{SR}$ is the predicted HR image, $I_{LR}$ is the input LR image, $\mathcal{C}_\omega(\cdot)$ is the first convolution parameterized by $\omega$,
$\mathcal{M}_\theta(\cdot)$ denotes stacked FMM modules parameterized by $\theta$, and $\mathcal{U}_\gamma(\cdot)$ represents the upsampler function parameterized by $\gamma$.
Following the previous work~\cite{ShuffleMixer}, these parameters are optimized using a combination of mean absolute error (MAE) loss and an FFT-based frequency loss function, which is defined as:
\begin{equation}
	\begin{split}
		\mathcal{L} =  \|I_{SR}-I_{HR}\|_1 + \lambda \|\mathcal{F}(I_{SR})-\mathcal{F}(I_{HR})\|_1,
	\end{split}
	\label{eqa:loss}
\end{equation}
where $I_{HR}$ is the high-quality ground-truth image, $ \|\cdot \|_1$ denotes the $L_1$-norm, $\mathcal{F}$ represents the Fast Fourier transform, and $\lambda$ is a weight parameter that is set to be 0.05 empirically.

\subsection{Spatially-adaptive Feature Modulation}
\label{sec:safm}
Compared to the self-attention mechanism~\cite{Swin, SwinIR, vit} or the large kernel convolution~\cite{replknet, van}, 
we propose a lightweight alternative to learn long-range dependencies from multi-scale feature representations so that more useful features can be better explored for HR image reconstruction.
As shown in Figure~\ref{fig:arc}, we apply a feature pyramid to generate an attention map for spatially-adaptive feature modulation.
To reduce the model complexity and obtain a pyramidal feature representation,
we first employ a channel split operation on the normalized input features, producing four-part components. 
A $3\times 3$ depth-wise convolution processes the first one, and the rest parts are fed into a multi-scale feature generation unit. 
Given the input feature $X$, this procedure can be expressed as: 
\begin{equation}
	\begin{split}
		& [X_0, X_1, X_2, X_3]  = \text{Split}(X),                 \\
		& \hat{X_0} = \text{DW-Conv}_{3\times 3}(X_0),                  \\
		& \hat{X_i} =  \uparrow_{p}(\text{DW-Conv}_{3\times 3}(\downarrow_{\frac{p}{2^i}}(X_i))), 1 \leq i \leq 3,            \\
	\end{split}
	\label{eqa:feats}
\end{equation}
where $\text{Split}(\cdot)$ is the channel split operation, 
$\text{DW-Conv}_{3\times 3}(\cdot)$ is a depth-wise convolution with a kernel size of $3\times 3$ pixels, 
$\uparrow_{p}(\cdot)$ represents upsampling features at a specific level to the original resolution $p$ via nearest interpolation for fast implementation, 
and $\downarrow_{\frac{p}{2^i}}$ denotes downsampling the input features to the size of $\frac{p}{2^i}$.
As we desire to select discriminative features towards learning non-local interactions, adaptive max pooling is applied over the input features to generate multi-scale features.
Results in Table~\ref{tab:ablation} illustrate that this max pooling operation helps to improve reconstruction performance.
We then concatenate these multi-scale features to aggregate local and global relations by a $1\times 1$ convolution. It can be expressed by:
\begin{equation}
	\begin{split}
		\hat{X} = \text{Conv}_{1\times 1}\left( \text{Concat}([\hat{X_0}, \hat{X_1}, \hat{X_2}, \hat{X_3}])\right),
	\end{split}
	\label{eqa:cat}
\end{equation}
where $\text{Concat}(\cdot)$ denotes a concatenation operation along the channel dimension, and $ \text{Conv}_{1\times 1}(\cdot)$ is the $1\times 1$ convolution.
After obtaining the refined representation $\hat{X}$, we normalize it through a GELU non-linearity~\cite{GELU} to estimate the attention map and adaptively modulate $X$ according to the estimated attention in an element-wise product.
This process can be written as: 
\begin{equation}
	\begin{split}
		\bar{X} = \phi(\hat{X}) \odot X,
	\end{split}
	\label{eqa:mod}
\end{equation}
where $\phi(\cdot)$ represents the GELU function and $\odot$ is the element-wise product.
Figure~\ref{fig:LAM} and Table~\ref{tab:runtime} intuitively illustrate that benefiting from the multi-scale feature representation, we can apply such a spatially-adaptive modulation mechanism for gathering long-range features with small memory and computational costs.
As shown in Table~\ref{tab:ablation}, this multi-scale form achieves better performance with less memory consumption than directly extracting features with a depth-wise convolution.
%

\subsection{Convolutional Channel Mixer}
\label{sec:ccm}
We note that the SAFM sub-block focuses on exploring the global information while the local contextual information also facilitates high-resolution image reconstruction.
Different from the commonly used feed-forward network~\cite{vit, Swin, SwinIR} that uses two consecutive $1\times 1$ convolutions to transform the features in channel dimensions for local context information exploration, 
we present a convolutional channel mixer (CCM) based on the FMBConv~\cite{EfficientNetv2} to enhance the local spatial modelling ability and perform channel mixing.
The proposed CCM contains a $3\times3$ convolution and a $1\times 1$ convolution. 
Within this, the first $3\times 3$ convolution encodes the spatially local contexts and \textbf{doubles} the number of channels of the input features for mixing channels; 
the later $1\times 1$ convolution reduces the channels back to the original input dimension.
A GELU~\cite{GELU} function is applied to the hidden layer for non-linear mapping.
This manner is more memory-efficient than employing a $3\times 3$ depth-wise convolution on the extended dimension (e.g. Inverted residual block~\cite{MobileNetV2}), as shown in Table~\ref{tab:ablation}.
Compared with the original FMBConv, we made the following modifications to make it more compatible with our architecture: (1) removing the squeeze-and-excitation (SE) block~\cite{SE}; (2) replacing the BatchNorm~\cite{BN} with the LayerNorm~\cite{LN} and moving it before the convolution.
Excluding the SE block mainly because the SAFM also has a dynamic utility on the channel dimension, and the reconstruction performance does not drop without it.
In addition, using the LayerNorm enables better stabilization of the model training and better results, as discussed in Section~\ref{sec:Any}.

\subsection{Feature Mixing Module}
\label{sec:fmm}
Motivated by the network design of ViT that contains a self-attention module for global feature aggregation and a feed-forward network for feature refinement, 
we formulate the proposed SAFM and the CCM into a unified feature mixing module to select representative features.
The feature mixing module can be formulated as: 
\begin{equation}
	\begin{split}
		& Y  = \text{SAFM}(\text{LN}(X)) + X,  \\
		& Z  = \text{CCM}(\text{LN}(Y)) + Y,  \\
	\end{split}
	\label{eq: fmm}
\end{equation}
where $\text{LN}(\cdot)$ is the LayerNorm~\cite{LN} layer, $X$, $Y$, and $Z$ are the intermediate features.
%
\begin{table*}[t]
	\caption{\textbf{Comparisons of efficient SR networks on the commonly used benchmark datasets.} 
		All PSNR/SSIM results are calculated on the \textbf{Y}-channel. 
		\#Acts represents all elements of the output of convolutional layers.
		\#FLOPs and \#Acts are measured corresponding to an HR image of the size $1280 \times 720$ pixels. 
		\textcolor{red}{Red} color denotes the best performance. 
		Blanked entries indicate results not reported or not available from previous work.
		$^\ast$ denotes that the results are obtained with the structural re-parameterization technique.
	}
	\label{tab:benchmark_result}
	\vspace{-5mm}
	\begin{center}
		\resizebox{\textwidth}{!}{
			\begin{tabular}{| l | c | c | c | c | c | c | c | c |c|}
				\hline
				Methods  & Scale & \#Params [K] & \#FLOPs [G]  & \#Acts [M]& Set5 & Set14 & B100 & Urban100 & Manga109 \\
				\hline
				Bicubic      &\multirow{16}*{$\times 2$} &-    &-      &-         &33.66/0.9299 &30.24/0.8688 &29.56/0.8431 &26.88/0.8403 &30.80/0.9339\\
				SRCNN~\cite{srcnn}      &~               &57     &53   &89        &36.66/0.9542 &32.42/0.9063 &31.36/0.8879 &29.50/0.8946 &35.74/0.9661 \\
				FSRCNN~\cite{FSRCNN}    &~               &12     &6    &41        &37.00/0.9558  &32.63/0.9088  &31.53/0.8920  &29.88/0.9020  &36.67/0.9694 \\
				ESPCN~\cite{ESPCN}      &~               &21     &5    &23        &36.83/0.9564  &32.40/0.9096  &31.29/0.8917  &29.48/0.8975  & - \\
				VDSR~\cite{VDSR}        &~               &665    &613  &1,120     &37.53/0.9587  &33.03/0.9124  &31.90/0.8960  &30.76/0.9140  &37.22/0.9729 \\
				LapSRN~\cite{LapSRN}    &~               &813    &30   &223       &37.52/0.9590  &33.08/0.9130  &31.80/0.8950  &30.41/0.9100  &37.27/0.9740 \\
				CARN-M~\cite{CARN}      &~               &415    &91   &655       &37.53/0.9583  &33.26/0.9141  &31.92/0.8960  &31.23/0.9193  & - \\
				CARN~\cite{CARN}        &~               &1,592  &223  &522       &37.76/0.9590  &33.52/0.9166  &32.09/0.8978  &31.92/0.9256  &- \\
				EDSR-baseline~\cite{EDSR} &~             &1,370  &316  &563       &37.99/0.9604  &33.57/0.9175  &32.16/0.8994  &31.98/0.9272  &38.54/0.9769 \\
				IMDN~\cite{IMDN}         &~              &694    &161  &423       &38.00/0.9605  &33.63/0.9177  &\textcolor{red}{32.19}/0.8996  &32.17/0.9283  &\textcolor{red}{38.88}/\textcolor{red}{0.9774} \\       
                PAN~\cite{PAN}          &~               &261    &71    &677      &38.00/0.9605  &33.59/0.9181 &32.18/0.8997 &32.01/0.9273 &38.70/0.9773\\
                LAPAR-A~\cite{LAPAR}    &~               &548    &171   &656  &\textcolor{red}{38.01}/0.9605  &33.62/\textcolor{red}{0.9183}  &\textcolor{red}{32.19}/\textcolor{red}{0.8999}  &32.10/0.9283 &38.67/0.9772 \\
				ECBSR-M16C64~\cite{ECBSR} &~             &596     &137     &252$^\ast$ &37.90/\textcolor{red}{0.9615}  &33.34/0.9178  &32.10/0.9018  &31.71/0.9250  &- \\
				SMSR~\cite{SMSR}          &~             &985     &132     &-          &38.00/0.9601  &\textcolor{red}{33.64}/0.9179  &32.17/0.8990  &\textcolor{red}{32.19}/\textcolor{red}{0.9284}  &38.76/0.9771 \\
				ShuffleMixer~\cite{ShuffleMixer} &~      &394     &91       &832    &\textcolor{red}{38.01}/0.9606  &33.63/0.9180  &32.17/0.8995  &31.89/0.9257  &38.83/\textcolor{red}{0.9774} \\
				\textbf{SAFMN}                    &~             &\textbf{228}    &\textbf{52}  &\textbf{299}  &38.00/0.9605  &33.54/0.9177  &32.16/0.8995  &31.84/0.9256  &38.71/0.9771\\
				
				\hline
				Bicubic             &\multirow{13}*{$\times 3$}     &-       &-      &-      &30.39/0.8682 &27.55/0.7742 &27.21/0.7385 &24.46/0.7349 &26.95/0.8556\\
				SRCNN~\cite{srcnn}   &~                             &57      &53     &89     &32.75/0.9090    &29.28/0.8209  &28.41/0.7863  &26.24/0.7989 &30.59/0.9107 \\
				FSRCNN~\cite{FSRCNN} &~                             &12      &5      &19     &33.16/0.9140 &29.43/0.8242 &28.53/0.7910  &26.43/0.8080  &30.98/0.9212 \\
				VDSR~\cite{VDSR}     &~                             &665     &613    &1,120  &33.66/0.9213 &29.77/0.8314 &28.82/0.7976  &27.14/0.8279  &32.01/0.9310 \\
				CARN-M~\cite{CARN}   &~                             &415     &46     &327    &33.99/0.9236  &30.08/0.8367  &28.91/0.8000  &27.55/0.8385  & - \\
				CARN~\cite{CARN}     &~                             &1,592    &119   &268    &34.29/0.9255  &30.29/0.8407  &29.06/0.8034  &28.06/0.8493  & - \\
				EDSR-baseline~\cite{EDSR} &~                        &1,555    &160   &285    &34.37/0.9270  &30.28/0.8417  &29.09/0.8052  &28.15/0.8527  &33.45/0.9439 \\
				IMDN~\cite{IMDN}      &~                            &703      &72    &190    &34.36/0.9270  &30.32/0.8417  &29.09/0.8046  &28.17/0.8519  &33.61/0.9445     \\ 
				PAN~\cite{PAN}        &~                            &261      &39    &340    &\textcolor{red}{34.40}/0.9271  &30.36/0.8423 &29.11/0.8050  &28.11/0.8511  &33.61/0.9448 \\
                LAPAR-A~\cite{LAPAR}  &~                            &594      &114   &505    &34.36/0.9267  &30.34/0.8421  &29.11/\textcolor{red}{0.8054}  &28.15/0.8523  &33.51/0.9441 \\
				SMSR~\cite{SMSR}      &~                            &993      &68    &-    &\textcolor{red}{34.40}/0.9270  &30.33/0.8412  &29.10/0.8050  &\textcolor{red}{28.25}/\textcolor{red}{0.8536}  &33.68/0.9445 \\
				ShuffleMixer~\cite{ShuffleMixer}  &~                &415     &43  &404 &\textcolor{red}{34.40}/\textcolor{red}{0.9272}  &\textcolor{red}{30.37}/\textcolor{red}{0.8423}  &\textcolor{red}{29.12}/0.8051  &28.08/0.8498  &\textcolor{red}{33.69}/\textcolor{red}{0.9448} \\
				\textbf{SAFMN}    &~  &\textbf{233}   &\textbf{23}  &\textbf{134}    &34.34/0.9267  &30.33/0.8418  &29.08/0.8048  &27.95/0.8474  &33.52/0.9437\\

				\hline
				Bicubic &\multirow{16}*{$\times 4$}       &-      &-     &-        &28.42/0.8104 &26.00/0.7027 &25.96/0.6675 &23.14/0.6577 &24.89/0.7866\\
				SRCNN~\cite{srcnn}            &~           &57    &53    &89       &30.48/0.8628  &27.49/0.7503   &26.90/0.7101  &24.52/0.7221 & 27.66/0.8505 \\
				FSRCNN~\cite{FSRCNN}          &~           &12    &5     &11       &30.71/0.8657  &27.59/0.7535  &26.98/0.7150   &24.62/0.7280  &27.90/0.8517 \\
				ESPCN~\cite{ESPCN}            &~           &25    &1     &6        &30.52/0.8697  &27.42/0.7606  &26.87/0.7216   &24.39/0.7241  &- \\
				VDSR~\cite{VDSR}              &~           &665   &613   &1,120    &31.35/0.8838  &28.01/0.7674  &27.29/0.7251   &25.18/0.7524  &28.83/0.8809 \\
				LapSRN~\cite{LapSRN}          &~           &813   &149   &264      &31.54/0.8850  &28.19/0.7720  &27.32/0.7280   &25.21/0.7560  &29.09/0.8845 \\
				CARN-M~\cite{CARN}            &~           &415   &33    &227      &31.92/0.8903  &28.42/0.7762  &27.44/0.7304   &25.62/0.7694  & - \\
				CARN~\cite{CARN}              &~           &1,592 &91    &194      &32.13/0.8937  &28.60/0.7806   &27.58/0.7349   &26.07/0.7837  & - \\
				EDSR-baseline~\cite{EDSR}     &~           &1,518 &114   &202      &32.09/0.8938  &28.58/0.7813  &27.57/0.7357   &26.04/0.7849  &30.35/0.9067 \\
				IMDN~\cite{IMDN}              &~           &715   &41    &108      &\textcolor{red}{32.21}/0.8948  &28.58/0.7811  &27.56/0.7353   &26.04/0.7838  &30.45/0.9075 \\ 
				PAN~\cite{PAN}                &~           &261   &22    &191      &32.13/0.8948 &28.61/0.7822 &27.59/0.7363 &26.11/0.7854 &30.51/0.9095\\
                LAPAR-A~\cite{LAPAR}          &~           &659   &94    &452      &32.15/0.8944  &28.61/0.7818  &\textcolor{red}{27.61}/\textcolor{red}{0.7366}   &\textcolor{red}{26.14}/\textcolor{red}{0.7871}  &30.42/0.9074 \\
				ECBSR-M16C64~\cite{ECBSR}     &~           &603    &35   &64$^\ast$   &31.92/0.8946  &28.34/0.7817 &27.48/0.7393   &25.81/0.7773  &- \\
				SMSR~\cite{SMSR}              &~           &1006   &42   &-           &32.12/0.8932  &28.55/0.7808  &27.55/0.7351   &26.11/0.7868  &30.54/0.9085\\ 	
				ShuffleMixer~\cite{ShuffleMixer}   &~      &411    &28   &269  &\textcolor{red}{32.21}/\textcolor{red}{0.8953}  &\textcolor{red}{28.66}/\textcolor{red}{0.7827}  &\textcolor{red}{27.61}/\textcolor{red}{0.7366}   &26.08/0.7835  &\textcolor{red}{30.65}/\textcolor{red}{0.9093} \\
				\textbf{SAFMN}                     &~       &\textbf{240}     &\textbf{14}   &\textbf{77}   &32.18/0.8948 &28.60/0.7813  &27.58/0.7359  &25.97/0.7809  &30.43/0.9063\\
				\hline
		\end{tabular}}
	\end{center}
	\vspace{-7mm}
\end{table*}

\section{Experimental Results}
In this section, we perform quantitative and qualitative evaluations to demonstrate the effectiveness of the proposed method. 
\label{sec:exp}
\subsection{Dataset and Implementation}
\noindent
\textbf{Datasets.} 
Following previous works~\cite{LAPAR, SwinIR, ShuffleMixer}, we use DIV2K~\cite{DIV2K} and Flickr2K~\cite{EDSR} as the training data and generate LR images by applying the bicubic downscaling to reference HR images. 
We use five commonly used benchmark datasets inluding Set5~\cite{Set5}, Set14~\cite{Set14}, B100~\cite{B100}, Urban100~\cite{Urban100} and Manga109~\cite{Manga109} as test data.
We use the peak signal to noise ratio (PSNR) and the structural similarity index (SSIM) to evaluate the quality of the restored images. 
All PSNR and SSIM values are calculated on the \textbf{Y} channel of images transformed to YCbCr color space.

\noindent
\textbf{Implementation details.}
During the training, the data argumentation is performed on the input patches with random horizontal flips and rotations. 
In addition, we randomly crop 64 patches of size $64\times64$ pixels from LR images as the basic training inputs. 
The number of FMM and feature channels is set to 8 and 36, respectively. 
We use the Adam~\cite{Adam} optimizer with $\beta_1=0.9$ and $\beta_2=0.99$ to solve the proposed model. The number of iterations is set to 500,000.  
We set the initial learning rate to $1\times10^{-3}$ and the minimum one to $1\times10^{-5}$, which is updated by the Cosine Annealing scheme~\cite{SGDR}. 
All experiments are conducted with the PyTorch framework on an NVIDIA GeForce RTX 3090 GPU. 
Due to the page limit, we include more results in the supplemental material. The training code and models will be available to the public.

\begin{figure*}
	\footnotesize
	\begin{center}
		\begin{tabular}{c c c c c c c c}
			\multicolumn{3}{c}{\multirow{5}*[35pt]{\includegraphics[width=0.35\linewidth, height=0.192\linewidth]{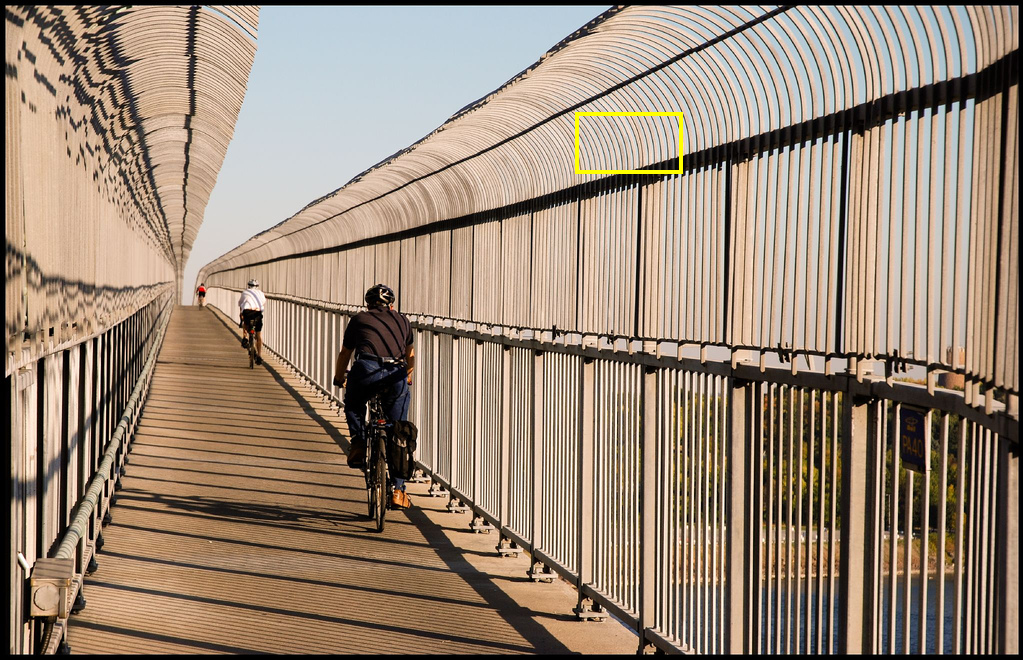}}}&\hspace{-3.0mm}
			\includegraphics[width=0.15\linewidth, height = 0.085\linewidth]{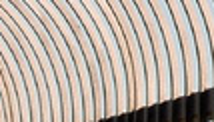} &\hspace{-3.0mm}
			\includegraphics[width=0.15\linewidth, height = 0.085\linewidth]{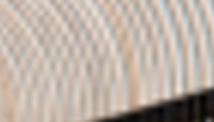}  &\hspace{-3.0mm}
			\includegraphics[width=0.15\linewidth, height = 0.085\linewidth]{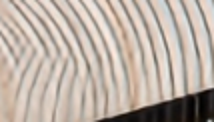}  &\hspace{-3.0mm}
			\includegraphics[width=0.15\linewidth, height = 0.085\linewidth]{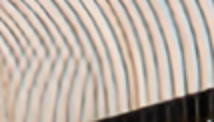} \\
			\multicolumn{3}{c}{~} &\hspace{-3.0mm}  (a) HR patch &\hspace{-3.0mm}  (b) Bicubic &\hspace{-3.0mm}  (c) VDSR~\cite{VDSR}  &\hspace{-3.0mm}  (d) ShuffleMixer~\cite{ShuffleMixer}\\
			
			\multicolumn{3}{c}{~} & \hspace{-3.0mm}
			\includegraphics[width=0.15\linewidth, height = 0.085\linewidth]{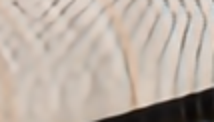} & \hspace{-3.0mm}
			\includegraphics[width=0.15\linewidth, height = 0.085\linewidth]{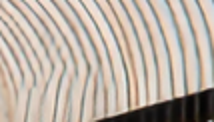} & \hspace{-3.0mm}
			\includegraphics[width=0.15\linewidth, height = 0.085\linewidth]{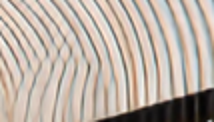} & \hspace{-3.0mm}
			\includegraphics[width=0.15\linewidth, height = 0.085\linewidth]{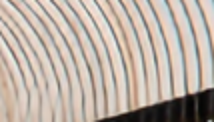} \\
			\multicolumn{3}{c}{\hspace{-3.0mm} img024 from Urban100} &\hspace{-3.0mm}  (e) LapSRN~\cite{LapSRN} &  \hspace{-3.0mm} (f) CARN~\cite{CARN}  &\hspace{-3.0mm}  (g) IMDN~\cite{IMDN}  & \hspace{-3.0mm}(h) SAFMN \\
			
			\\
			
			\multicolumn{3}{c}{\multirow{5}*[35pt]{\includegraphics[width=0.35\linewidth, height=0.192\linewidth]{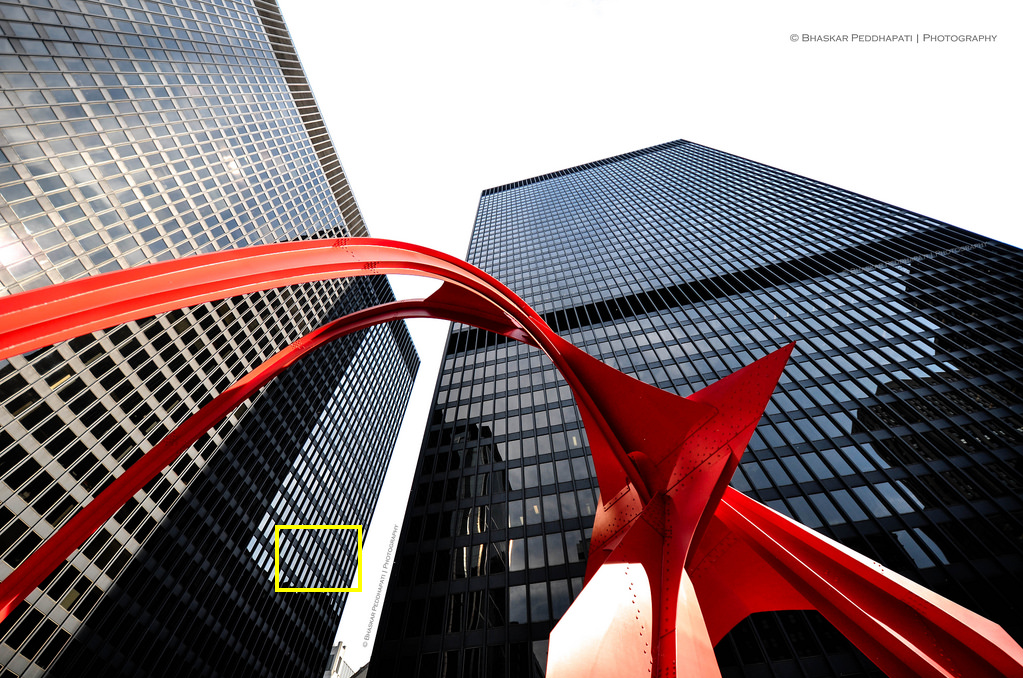}}} &\hspace{-3.0mm}
			\includegraphics[width=0.15\linewidth, height = 0.085\linewidth]{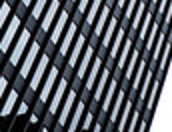} &\hspace{-3.0mm}
			\includegraphics[width=0.15\linewidth, height = 0.085\linewidth]{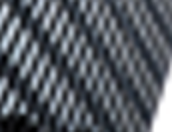} &\hspace{-3.0mm}
			\includegraphics[width=0.15\linewidth, height = 0.085\linewidth]{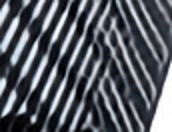} &\hspace{-3.0mm}
			\includegraphics[width=0.15\linewidth, height = 0.085\linewidth]{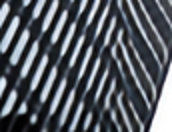} \\
			\multicolumn{3}{c}{~} &\hspace{-3.0mm} (a) HR patch &\hspace{-3.0mm} (b) Bicubic &\hspace{-3.0mm}  (c) VDSR~\cite{VDSR}  &\hspace{-3.0mm} (d)  ShuffleMixer~\cite{ShuffleMixer}\\
			
			\multicolumn{3}{c}{~} & \hspace{-3.0mm}
			\includegraphics[width=0.15\linewidth, height = 0.085\linewidth]{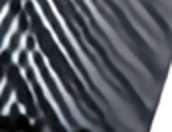} & \hspace{-3.0mm}
			\includegraphics[width=0.15\linewidth, height = 0.085\linewidth]{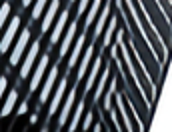}   & \hspace{-3.0mm}
			\includegraphics[width=0.15\linewidth, height = 0.085\linewidth]{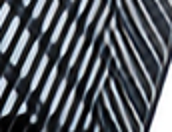} & \hspace{-3.0mm}
			\includegraphics[width=0.15\linewidth, height = 0.085\linewidth]{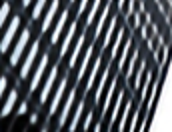} \\
			\multicolumn{3}{c}{img062 from Urban100} &\hspace{-3.0mm} (e) LapSRN~\cite{LapSRN} &  \hspace{-3.0mm} (f) CARN~\cite{CARN}  &\hspace{-3.0mm}  (g) IMDN~\cite{IMDN}  & \hspace{-3.0mm} (h) SAFMN\\
			
			\\
			\multicolumn{3}{c}{\multirow{5}*[35pt]{\includegraphics[width=0.35\linewidth, height=0.192\linewidth]{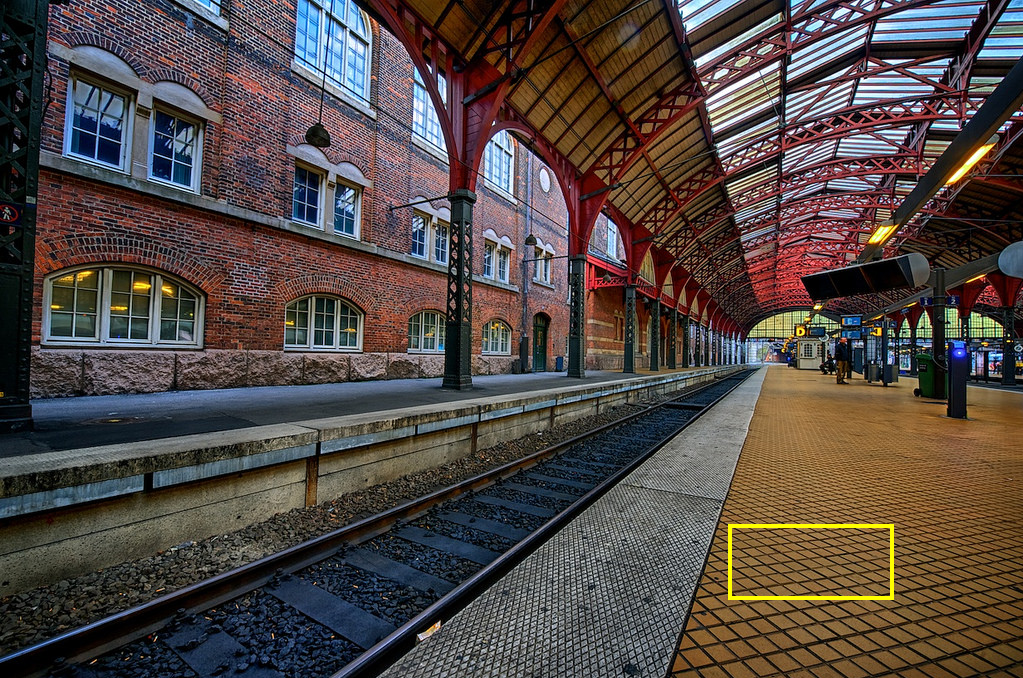}}} &\hspace{-3.0mm}
			\includegraphics[width=0.15\linewidth, height = 0.085\linewidth]{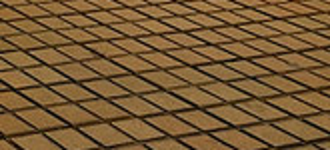} &\hspace{-3.0mm}
			\includegraphics[width=0.15\linewidth, height = 0.085\linewidth]{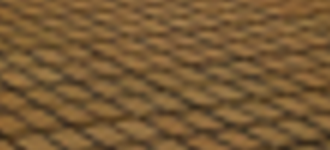} &\hspace{-3.0mm}
			\includegraphics[width=0.15\linewidth, height = 0.085\linewidth]{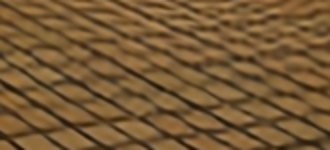} &\hspace{-3.0mm}
			\includegraphics[width=0.15\linewidth, height = 0.085\linewidth]{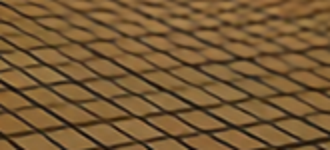} \\
			\multicolumn{3}{c}{~} &\hspace{-3.0mm} (a) HR patch &\hspace{-3.0mm} (b) Bicubic &\hspace{-3.0mm} (c) VDSR~\cite{VDSR}  &\hspace{-3.0mm} (d)  ShuffleMixer~\cite{ShuffleMixer}\\
			
			\multicolumn{3}{c}{~} & \hspace{-3.0mm}
			\includegraphics[width=0.15\linewidth, height = 0.085\linewidth]{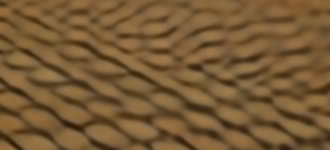} & \hspace{-3.0mm}
			\includegraphics[width=0.15\linewidth, height = 0.085\linewidth]{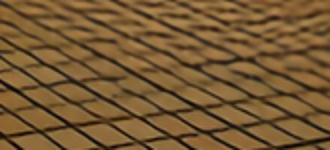} & \hspace{-3.0mm}
			\includegraphics[width=0.15\linewidth, height = 0.085\linewidth]{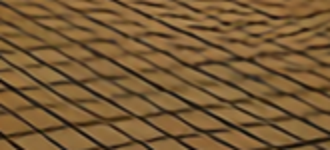} & \hspace{-3.0mm}
			\includegraphics[width=0.15\linewidth, height = 0.085\linewidth]{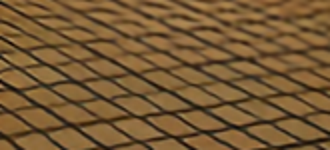} \\
			\multicolumn{3}{c}{img098 from Urban100} &\hspace{-3.0mm} (e) LapSRN~\cite{LapSRN} & \hspace{-3.0mm} (f) CARN~\cite{CARN}  &\hspace{-3.0mm}  (g) IMDN~\cite{IMDN}  & \hspace{-3.0mm} (h) SAFMN\\			
		\end{tabular}
	\end{center}
	\vspace{-3mm}
	\caption{Visual comparisons for $\times3$ SR on the Urban100 dataset. The proposed method recovers the image with clearer structures.}
	\label{fig:visual_results}
	\vspace{-5mm}
\end{figure*}

\subsection{Comparisons with State-of-the-Art Methods}
\noindent
\textbf{Quantitative comparisons.}
To evaluate the performance of our approach, we compare it with state-of-the-art lightweight SR methods, including SRCNN~\cite{srcnn}, FSRCNN~\cite{FSRCNN}, ESPCN~\cite{ESPCN}, VDSR~\cite{VDSR}, LapSRN~\cite{LapSRN}, CARN~\cite{CARN}, EDSR-baseline~\cite{EDSR}, IMDN~\cite{IMDN}, PAN~\cite{PAN}, LAPAR~\cite{LAPAR}, ECBSR~\cite{ECBSR}, SMSR~\cite{SMSR}, and ShuffleMixer~\cite{ShuffleMixer}.

The quantitative comparisons on benchmark datasets for the upscaling factors of $\times 2$, $\times 3$, and $\times 4$ are reported in Table~\ref{tab:benchmark_result}.
In addition to PSNR/SSIM metrics, we list the number of parameters (\#Params), FLOPs (\#FLOPs) and activations (\#Acts). 
We calculate the number of FLOPs and activations with the fvcore\footnote{We use the \href{https://detectron2.readthedocs.io/en/latest/modules/fvcore.html\#fvcore.nn.parameter\_count}{fvcore.nn.flop\_count\_str} command to calculate the number of parameters, FLOPs and activations.} library under a setting of super-resolving an LR image to $1280 \times 720$ pixels. 
Among these metrics, \#Params and \#Acts are linked to memory consumption, and \#FLOPs is related to energy usage.
In particular, \#Acts is a better metric for measuring the efficiency of a model than the number of parameters and FLOPs as suggested in recent works~\cite{DNDS, AIM19constrainedSR, aim_efficientsr, ntire22efficientsr}.

Benefiting from the simple yet efficient structure, the proposed SAFMN obtains comparable performance with significantly fewer parameters and memory consumption.
Take $\times 4$ SR on B100 dataset as an example, our SAFMN has parameters about 85\% less than the CARN~\cite{CARN}, 66\% less than the IMDN~\cite{IMDN}, and 42\% less than the ShuffleMixer~\cite{ShuffleMixer}.
As for activations, we have 60\%, 29\% and 71\% fewer than them, respectively. 
While our model has a smaller footprint, we achieve similar performance among these methods.
Moreover, we compare the reconstruction accuracy, FLOPs and parameters on the $\times2$ Set5 dataset in Figure~\ref{fig:comp}. 
The proposed SAFMN model achieves a favorable trade-off between model complexity and reconstruction performance.

\noindent
\textbf{Qualitative comparisons.}
In addition to the quantitative evaluations, we provide qualitative comparisons of the proposed method. Figure~\ref{fig:visual_results} shows the visual comparisons on the Urban100 dataset for $\times 3$ SR. 
Our approach generates parallel straight lines and grid patterns more accurately than the listed methods.
These results also demonstrate the effectiveness of our method for adaptive feature modulation by exploiting non-local feature interactions.

\noindent
\textbf{Memory and running time comparisons.}
To fully examine the efficiency of our proposed method, 
we further evaluate our method against five representative ones, including 
CARN-M~\cite{CARN}, CARN~\cite{CARN}, EDSR-baseline~\cite{EDSR}, IMDN~\cite{IMDN},  and LAPAR-A~\cite{LAPAR} on $\times 4$ SR in terms of the GPU memory consumption (\#GPU Mem.) and running time (\#Avg. Time). 
The maximum GPU memory consumption is recorded during the inference.  
The running time is averaged on 50 test images with a $320 \times 180$ resolution.
We show the memory and running time comparisons in Table~\ref{tab:runtime},
our method achieves a clear improvement over other state-of-the-art methods.  
By using the multi-scale modulation layer and the efficient channel mixer, 
the GPU consumption of our SAFMN is only 10\% of the CARN series and 4\% of LAPAR-A; 
the running time is nearly twice as fast as other evaluated methods, except for IMDN.
Compared to IMDN~\cite{IMDN}, our method has a similar running time speed while significantly reducing memory usage.
Tables~\ref{tab:benchmark_result} and~\ref{tab:runtime} show that the proposed model achieves a favorable trade-off in terms of inference speed, model complexity and reconstruction performance against state-of-the-art methods.

\begin{table}
	\caption{\textbf{Memory and running time comparisons on $\times 4$ SR.}
		\#GPU Mem. denotes the maximum GPU memory consumption during the inference phase, 
		derived with the Pytorch \href{https://pytorch.org/docs/stable/generated/torch.cuda.max\_memory\_allocated.html}{torch.cuda.max\_memory\_allocated()} function. 
		\#Avg. Time is the average running time on 50 LR images with a size of $320\times180$ pixels.}
	\vspace{-2mm}
	\centering
	\small
	\begin{tabular}{|l|c|c|}
		\hline
		Methods    &\#GPU Mem. [M]   &\#Avg.Time [ms]\\
		\hline
		CARN-M~\cite{CARN}                  & 680.84  & 17.85   \\ 
		CARN~\cite{CARN}                      & 689.83  & 18.90    \\ 
		EDSR-baseline~\cite{EDSR}          & 486.58  & 19.81     \\ 
		IMDN~\cite{IMDN}                       & 203.44  & 10.22      \\ 
		LAPAR-A~\cite{LAPAR}              & 1811.47 & 24.91      \\ 
		SAFMN                                         & 65.26   & 10.71     \\ 
		\hline
	\end{tabular}
	\label{tab:runtime}
	\vspace{-3mm}
\end{table}

\begin{table*}
	\caption{\textbf{Ablation for SAFMN on DIV2K-val and Manga109 datasets.} 
		SAFMN with a scaling factor of $\times 4$ is utilized as the baseline for ablation studies.
		The PSNR/SSIM values on benchmarks are reported. 
		``A $\rightarrow$ B'' is to replace A with B.
		``None'' means to remove the operation.
		``lr'' denotes the learning rate.
		``FBN'' is the abbreviation of the Frozen BatchNorm.
		``$\text{L}_2$ normalization'' represents that the inputs are normalized by $\text{L}_2$-norm over the channel dimension.
		``FM'', ``MR'', and ``FA'' are the abbreviations corresponding to feature modulation, multi-scale representation, and feature aggregation.
		$^\ast$ indicates that the corresponding results are obtained before the training collapse. 
		The numbers of parameters, \#FLOPs and \#Acts are counted by the fvcore library with a resolution of $320 \times 180$ pixels.}
	\centering
	\vspace{-1mm}
	\resizebox{\textwidth}{!}{
		\begin{tabular}{|c|l|c|c|c|c|c|}
			\hline
			Ablation  &Variant &\#Params [K] &\#FLOPs [G] &\#Acts [M] &DIV2K-val &Manga109\\
			\hline
			\textbf{Baseline}   & -          & \textbf{239.52}  & \textbf{13.56}  & \textbf{76.70}   & \textbf{30.43}/\textbf{0.8372}  & \textbf{30.43}/\textbf{0.9063}\\ 
			\hline
			\multirow{2}{*}{Main module}
			& SAFM $\rightarrow$ None                                             & 225.41  & 12.90  & 54.61   & 30.26/0.8330  &30.09/0.9018\\ 
			& CCM $\rightarrow$ None                                              & 30.72   & 1.61   & 26.93   & 29.69/0.8193  &28.49/0.8193 \\ 
			\hline
			\multirow{10}{*}{SAFM}
			& (a): w/o FM                                                                     & 239.52  & 13.56  & 76.70   & 30.36/0.8357  & 30.32/0.9048\\ 
			& (b): w/o MR                                                                    & 239.52  & 13.64  & 87.78   & 30.34/0.8350  & 30.30/0.9047\\ 
			& (c): w/o FA                                                                      & 228.86  & 12.96  & 60.11   & 30.36/0.8355  & 30.29/0.9049\\ 
			& (a) + (b)                                                                           & 239.52  & 13.64  & 87.78   & 30.32/0.8345  & 30.24/0.9038\\ 
			& (a) + (c)                                                                           & 228.86  & 12.96  & 60.11   & 30.34/0.8351  & 30.28/0.9043\\ 
			& (a) + (b) + (c)                                                                   & 228.86  & 13.05  & 71.19   & 30.31/0.8344  & 30.23/0.9036\\ 
			& AdaptiveMaxPool $\rightarrow$ AdaptiveAvgPool             & 239.52  & 13.56  & 76.70   & 30.40/0.8364  & 30.40/0.9061\\ 
			& AdaptiveMaxPool $\rightarrow$ Nearest interpolate            & 239.52  & 13.56  & 76.70   & 30.36/0.8354  & 30.31/0.9048\\ 
			& GELU $\rightarrow$ None                                                 & 239.52  & 13.56  & 76.70   & 30.40/0.8366  & 30.37/0.9058\\
			& GELU $\rightarrow$ Sigmoid                                             & 239.52  & 13.56  & 76.70   & 30.35/0.8355  & 30.29/0.9044\\
			\hline
			\multirow{3}{*}{CCM}
			& w/ SE                                                                                  & 260.98  & 13.59  & 76.70   & 30.39/0.8360  & 30.46/0.9067\\ 
			& CCM $\rightarrow$ Channel MLP                                         & 73.63   & 4.00   & 76.70   & 30.17/0.8313  & 29.80/0.8980\\ 
			& CCM $\rightarrow$ Inverted residual block                           & 245.28  & 13.85  & 110.00  & 30.43/0.8373  & 30.43/0.9064\\
			\hline
			\multirow{5}{*}{Normalization}
			& LN $\rightarrow$ None, lr=$1\times10^{-3}$$^\ast$                   & 238.37  & 13.55  & 76.70   & 30.29/0.8340  & 30.04/0.9014\\ 
			& LN $\rightarrow$ None, lr=$1\times10^{-4}$                             & 238.37  & 13.55  & 76.70   & 30.15/0.8306  & 29.74/0.8970\\
			& LN $\rightarrow$ BN                                                                 & 239.52  & 13.72  & 76.70   & 30.28/0.8354  & 30.05/0.9029\\ 
			& LN $\rightarrow$ FBN$^\ast$                                             & 238.37  & 13.55  & 76.70   & 30.30/0.8343  & 30.15/0.9028 \\
			& LN $\rightarrow$ $\text{L}_2$ normalization                               & 238.37  & 13.55  & 76.70   & 30.39/0.8358  & 30.31/0.9049 \\ 
			\hline
	\end{tabular}}
	\smallskip
	\label{tab:ablation}
	\vspace{-4mm}
\end{table*}

\section{Analysis and Discussion}
\label{sec:Any}
We further conduct extensive ablation studies to better understand and evaluate each component in the proposed SAFMN. 
For fair comparisons with the designed baselines, we implement all experiments based on $\times 4$ SAFMN and train them with the same setting. 
The experimental results in Table~\ref{tab:ablation} are measured on DIV2K-val~\cite{DIV2K} and Manga109~\cite{Manga109} datasets.

\noindent
\textbf{Effectiveness of the spatially-adaptive feature modulation.}
To demonstrate the effect of the spatially-adaptive feature modulation, we first remove this module for comparison.
Without it, the PSNR values will drop by 0.17dB (30.26 vs. 30.43) and 0.34dB (30.09 vs. 30.43) on the DIV2K-val and Manga109 datasets, respectively.
These results show the importance of the SAFM.
Therefore, we further investigate this module to find out why it works.
\begin{itemize}
	\vspace{-1mm}
	\item \textbf{Feature modulation.} 
	The feature modulation mechanism is introduced to enable the network with adaptive properties.
	Without this operation, the baseline model is lowered by 0.11dB on the Manga109 dataset.
	\vspace{-1mm}
	\item \textbf{Multi-scale representation.} 
	Here, ``w/o MR'' in Table~\ref{tab:ablation} denotes that we directly use a depth-wise convolution with a kernel size of $3 \times 3$ pixels to extract spatial information and do not employ multi-scale features.
	A noticeable performance drop of 0.13dB on the Manga109 dataset is observed without these multi-scale features.
	Moreover, we apply the adaptive max pooling over the input features in this module to build feature pyramids.
	Compared to using adaptive average pooling or nearest interpolation, 
	adaptive max pooling allows the model to detect discriminative features, resulting in better reconstruction results.
	
	\vspace{-1mm}
	\item \textbf{Feature aggregation.} 
	We use a $1\times 1$ convolution to aggregate the multi-scale features on the channel dimension.  
	Combined with the modulation mechanism, it brings a PSNR improvement of 0.14dB on the Manga109 dataset, which proves the necessity of aggregating the multi-scale features.

	Without all three components mentioned before, 
	it represents that only a $3\times 3$ depth-wise convolution is used to encode the spatial information 
	and leads to a PSNR reduction of 0.12dB and 0.2dB on the DIV2K-val and Manga109 datasets, respectively.
	This performance drop suggests that the spatially-adaptive modulation based on the multi-scale feature representation effectively boosts SR reconstruction performance.

	\vspace{-1mm}
	\item \textbf{GELU function.} 
	Here, we use the GELU~\cite{GELU} function to normalize the modulation map. 
	Table~\ref{tab:ablation} shows that better experimental results can be achieved with GELU than with Sigmoid or without GELU, 
	mainly because it weights the input features by their percentile and allows better-activated features.
\end{itemize}

\noindent
The above analysis shows that benefiting from the multi-scale feature representation, the proposed SAFM can effectively exploit long-range interactions.

\noindent
\textbf{Effectiveness of the convolutional channel mixer.}
Compared with the original FMBConv~\cite{EfficientNetv2}, the main change made by CCM is removing the SE~\cite{SE} block.
As shown in Table~\ref{tab:ablation}, using SE blocks, 
its performance is almost the same as without them. 
This result is mainly caused by the SAFM block already performing dynamic channel-wise feature recalibration.
Hence, we do not use the SE blocks for saving parameters.
We next conduct a series of ablations to verify that CCM can effectively encode local contextual information and perform channel mixing.
Without it, the model only achieves the accuracy of 29.69dB and 28.49dB on DIV2K-val and Manga109 datasets, proving the indispensability and locality modelling ability of this part.
We then change CCM to other channel mixers commonly used in ViT architectures: channel MLP~\cite{vit} or inverted residual block~\cite{MobileNetV2}. 
When the channel MLP is adopted for the channel mixer, there is a significant performance drop of 0.63dB (29.80 vs. 30.43) on the Manga109 dataset. 
%
\begin{figure}[thbp] 
	\centering
	\includegraphics[width=0.23\textwidth]{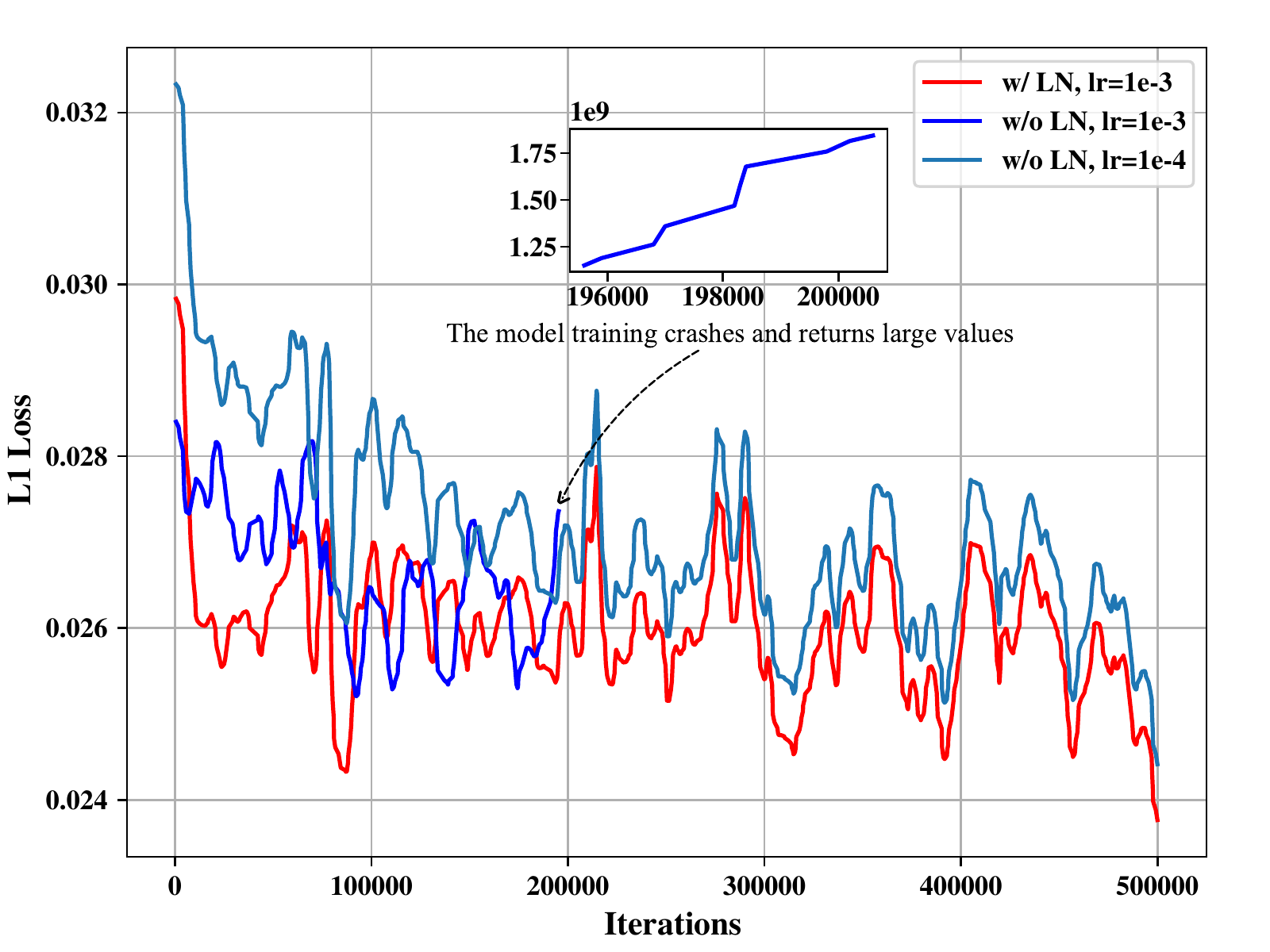}
	\includegraphics[width=0.24\textwidth]{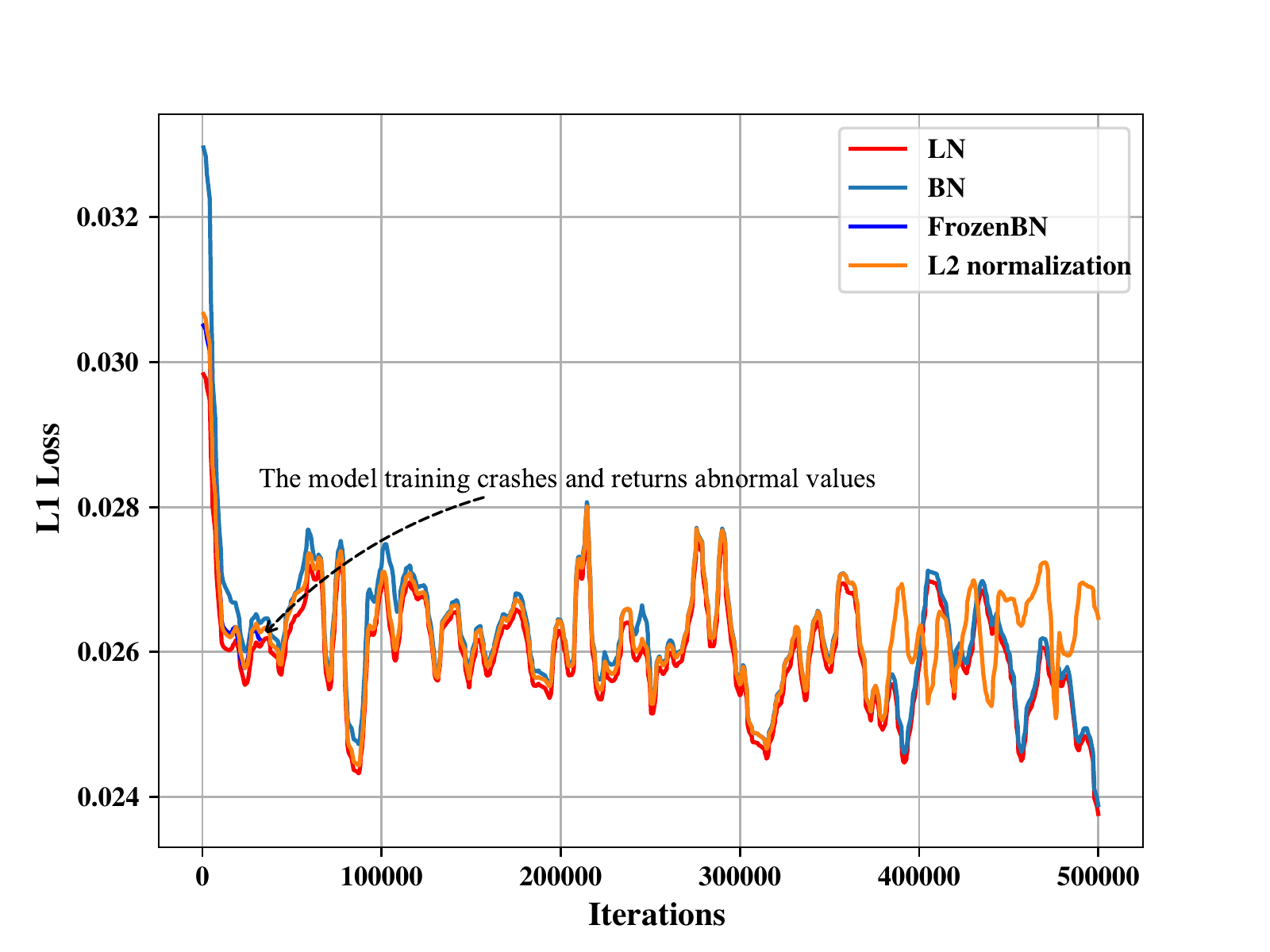}
	\\
	\makebox[0.23\textwidth]{\footnotesize (a) Effect of LayerNorm}
	\makebox[0.24\textwidth]{\footnotesize (b) Comparison of different normalizations}
	\vspace{-3mm}
	\caption{Smoothed training loss curves with or without normalization. Figure (a) shows LayerNorm stabilizes model training and converges better than without it. Figure (b) compares the effects of different normalization approaches on model training.}
	\label{fig:norm}
	\vspace{-5mm}
\end{figure}
This performance drop results from the lack of local spatial modelling ability.
For the inverted residual block, the performance is highly similar to that of CCM, 
but the corresponding \#Acts is increased by nearly 35M, which means more memory consumption and slower inference time.
Thus, we use the CCM as the default channel mixer.

\noindent
\textbf{Effectiveness of the LayerNorm layer.}
Since we use the element-wise product in the SAFM module, it will result in abnormal gradient values and unstable model training, as shown in Figure~\ref{fig:norm}(a).
It is, therefore, necessary to normalize the input features.
Here, we perform normalization with the LayerNorm~\cite{LN} layer. 
To verify this assumption, we first remove the LayerNorm layer.
Table~\ref{tab:ablation} and Figure~\ref{fig:norm}(a) suggest that without it, the model suffers a training crash at a large learning rate (i.e., $1\times10^{-3}$) and does not converge well at small ones (i.e., $1\times10^{-4}$) with its PSNR of only 30.15dB and 29.74dB on the DIV2K-val and Manga109 datasets, respectively.
Next, we compare the LayerNorm with three representative normalization methods, including BatchNorm~\cite{BN}, Frozen BatchNorm (FBN)~\cite{BN}, and $\text{L}_2$ normalization in Figure~\ref{fig:norm}(b).
The results in Table~\ref{tab:ablation} and Figure~\ref{fig:norm}(b) demonstrate that the BN family decrease the PSNR/SSIM values by a large margin, and the FBN even does not guarantee a stable model training process.
Although the $\text{L}_2$ normalization allows the model to be trained successfully, its performance is not as good as that of using LayerNorm.
Thus, the LayerNorm layer is set as default for the proposed SAFMN.
\section{Conclution}
\label{sec:con}
In this paper, we propose a simple yet efficient deep CNN model to solve the efficient image super-resolution problem.
The proposed SAFMN explores the  long-range adaptability  upon a  multi-scale feature representation-based modulation mechanism.
To complement the local contextual information, we further develop a compact  convolutional channel mixer to encode spatially local context and conduct channel mixing simultaneously.
We both qualitatively and quantitatively evaluate the proposed method on commonly used benchmarks. 
Experimental results show that the proposed SAFMN model is more efficient than state-of-the-art methods while achieving competitive performance.

{\small
	\bibliographystyle{ieee_fullname}
	\bibliography{egbib}
}

\end{document}


\title{Spatially-Adaptive Feature Modulation for Efficient Image Super-Resolution\\ - Supplemental Material -}
	
\author{~~~~Long Sun, Jiangxin Dong, Jinhui Tang, Jinshan Pan~~~~~\\
	~~~~Nanjing University of Science and Technology~~~~\\}
\maketitle

\section*{\centering Overview}
In this document, we further demonstrate the effectiveness of the proposed spatially-adaptive feature modulation and the LayerNorm layer in Section~\ref{sec: ablation}.
Then, we evaluate our method with the challenge winners in Section~\ref{sec: challenge}. We further compare the proposed method with classic performance-oriented SR models in Section~\ref{sec: classic_sr}.
Next, we make some notes on the Urban100 dataset in Section~\ref{sec: urban100_notes}.
Finally, we show more visual comparisons in Section~\ref{sec: visual}.

\section{Ablations of the spatially-adaptive feature modulation and the LayerNorm}
\label{sec: ablation}
\begin{figure*}
    \centering
    \includegraphics[width=0.155\linewidth]{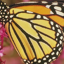}
    \includegraphics[width=0.16\linewidth]{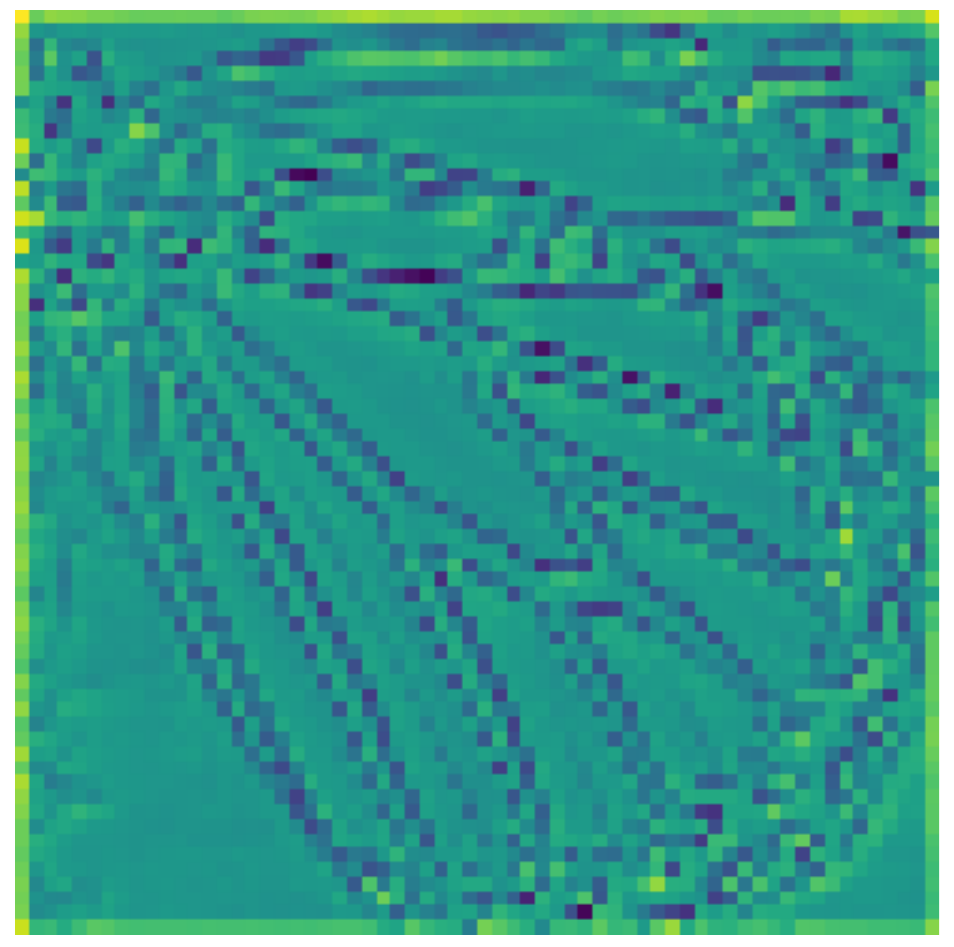}
    \includegraphics[width=0.16\linewidth]{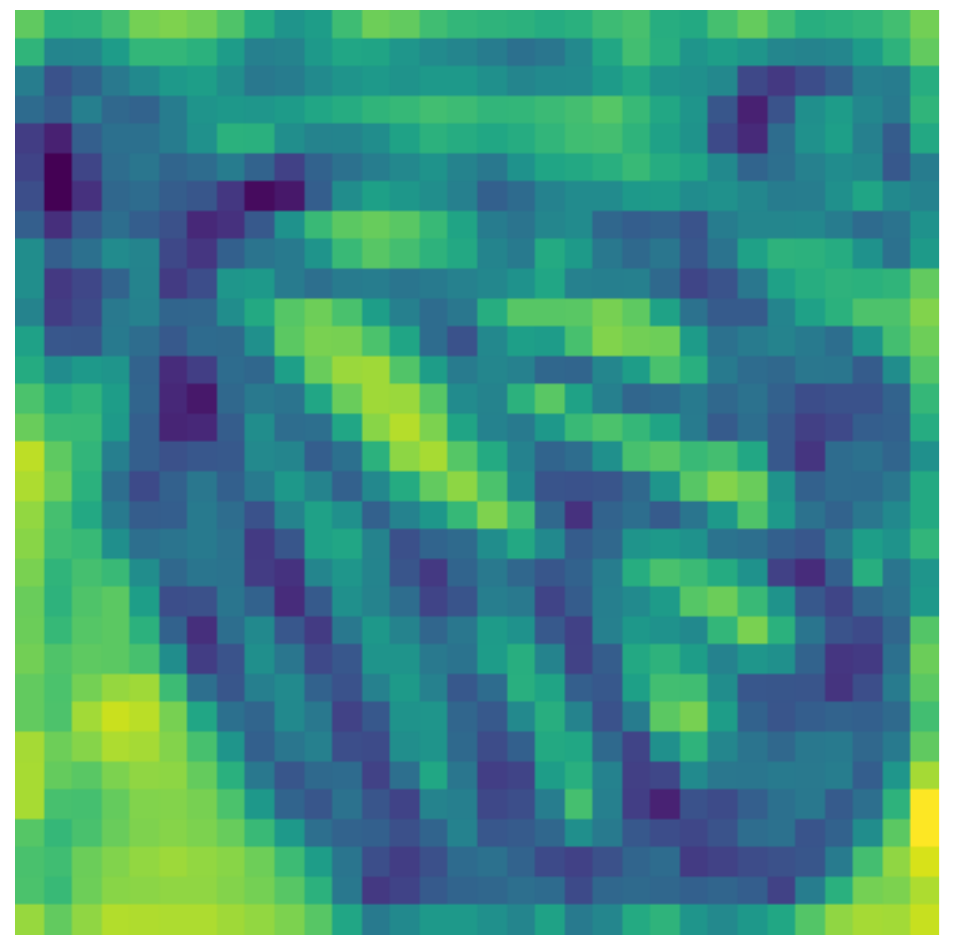}
    \includegraphics[width=0.16\linewidth]{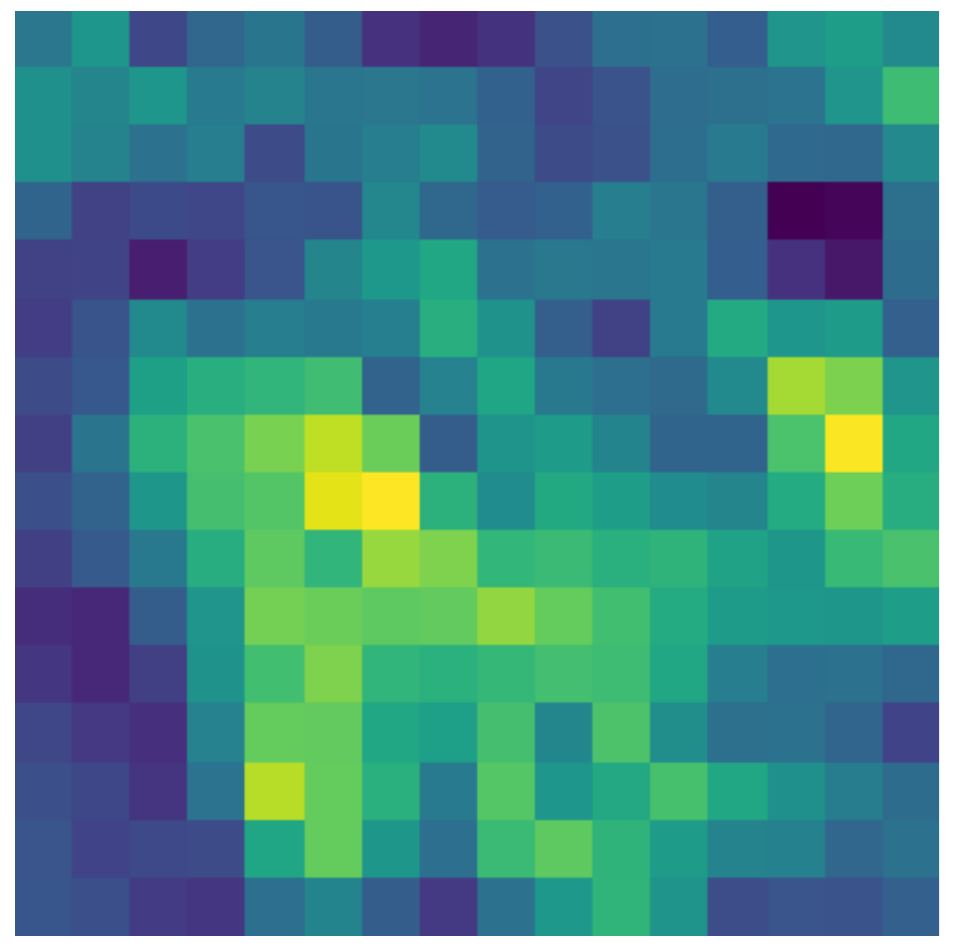}
    \includegraphics[width=0.16\linewidth]{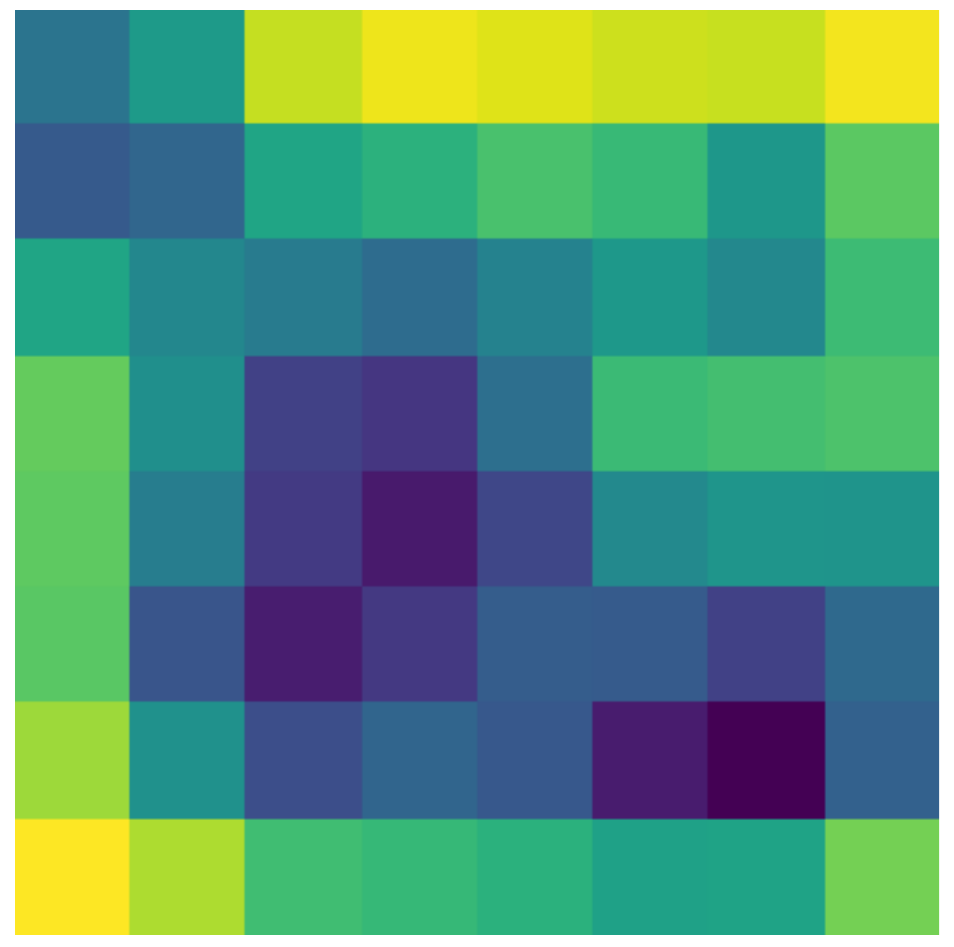}
    \includegraphics[width=0.16\linewidth]{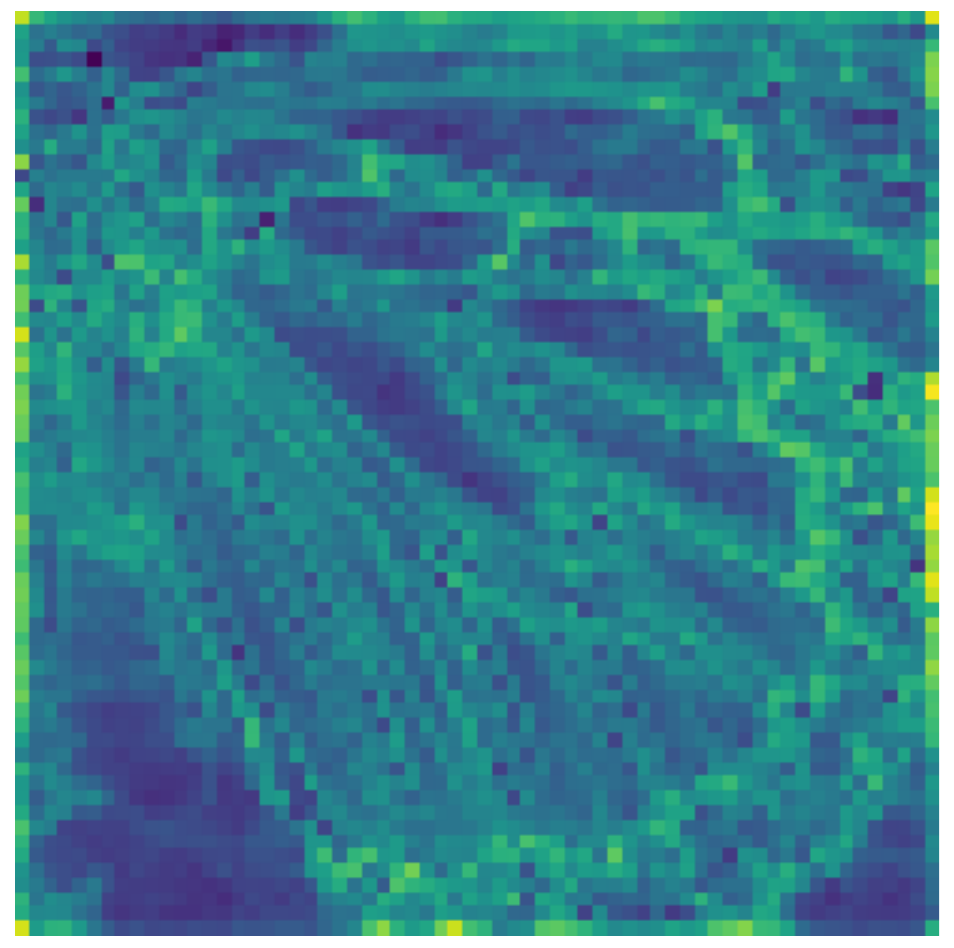}
    \\
    \makebox[0.155\linewidth]{\small (a) LR input}
    \makebox[0.16\linewidth]{\small (b) Scale 1}
    \makebox[0.16\linewidth]{\small (c) Scale 2}
    \makebox[0.16\linewidth]{\small (d) Scale 4}
    \makebox[0.16\linewidth]{\small (e) Scale 8}
    \makebox[0.16\linewidth]{\small (f) Aggregated feature}
    \\
    \vspace{-3mm}
    \caption{Illustration of learned deep features from the SAFM at different scales.}
    \label{fig:Scale}
    \vspace{-2mm}
\end{figure*}

\noindent
\textbf{Effect of scales in the spatially-adaptive feature modulation.}
We evaluate the effect of features at different scales in the spatially-adaptive feature modulation (SAFM) on the $\times 4$ DIV2K validation set.
Table~\ref{tab:Scale} shows that removing any scale information affects the reconstruction performance.
We further show the visualization of learned features at different scales to better understand why the SAFM works.
Figure~\ref{fig:Scale} illustrates that the learned features are independent and complementary between scales, and aggregating these features helps grab the overall structure.

\begin{figure*}[t] \centering
    \begin{minipage}{0.1\textwidth} \centering
        \includegraphics[width=1.2\linewidth]{figs/SAFM/butterfly_x4LR.png}\\
        \makebox[\linewidth]{\small (a) LR input}
    \end{minipage}
    \hspace{3mm}
    \begin{minipage}{0.85\textwidth}%
        \includegraphics[width=0.23\linewidth]{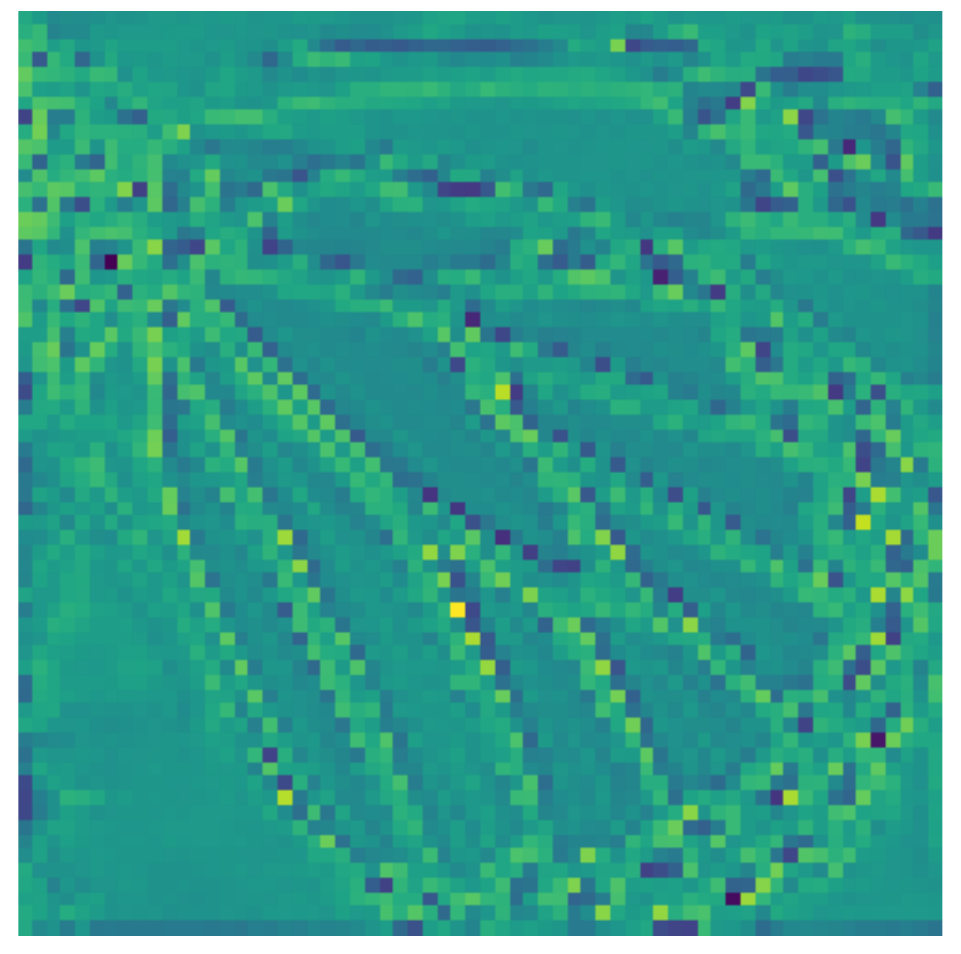}
        \includegraphics[width=0.23\linewidth]{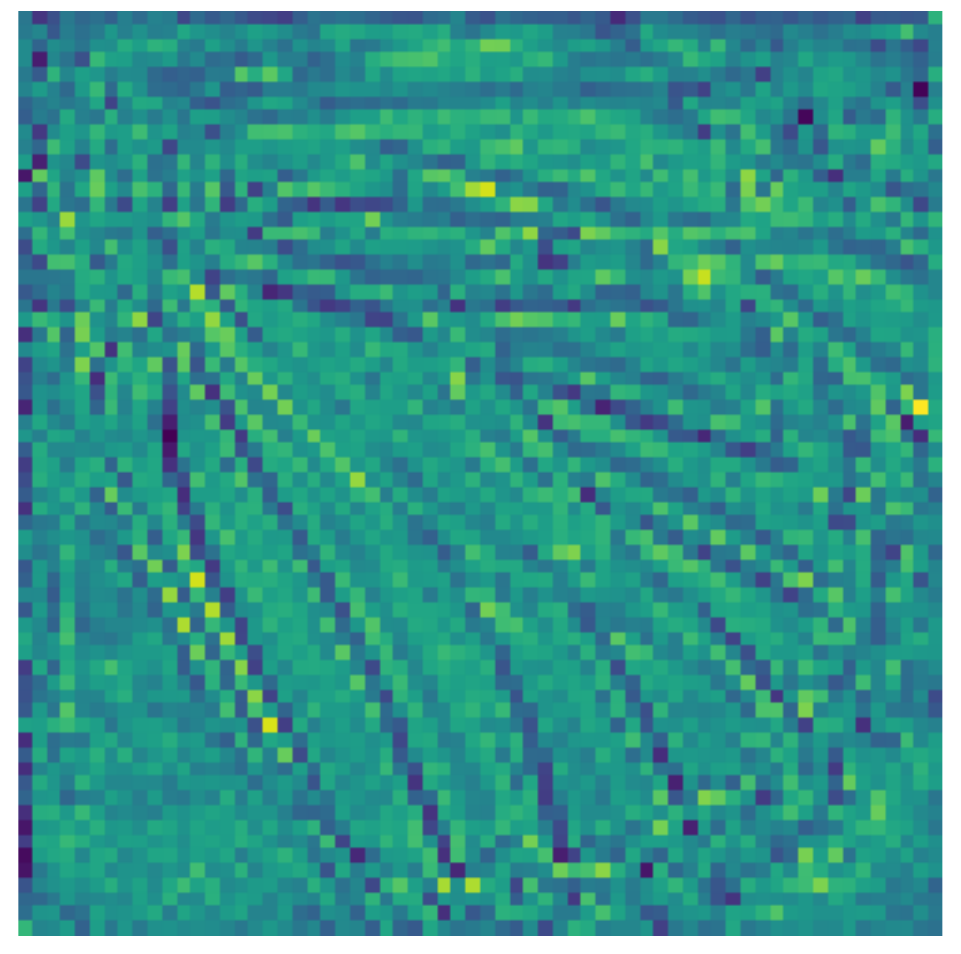}
        \includegraphics[width=0.23\linewidth]{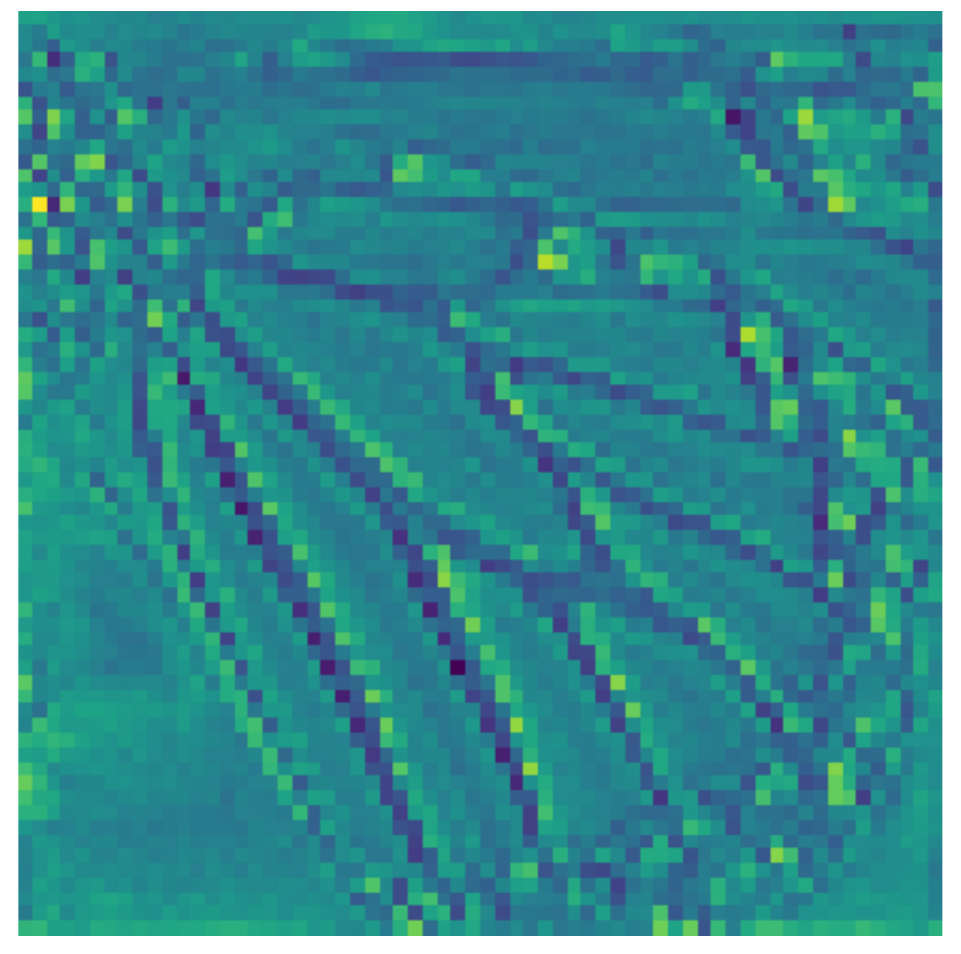}
        \includegraphics[width=0.23\linewidth]{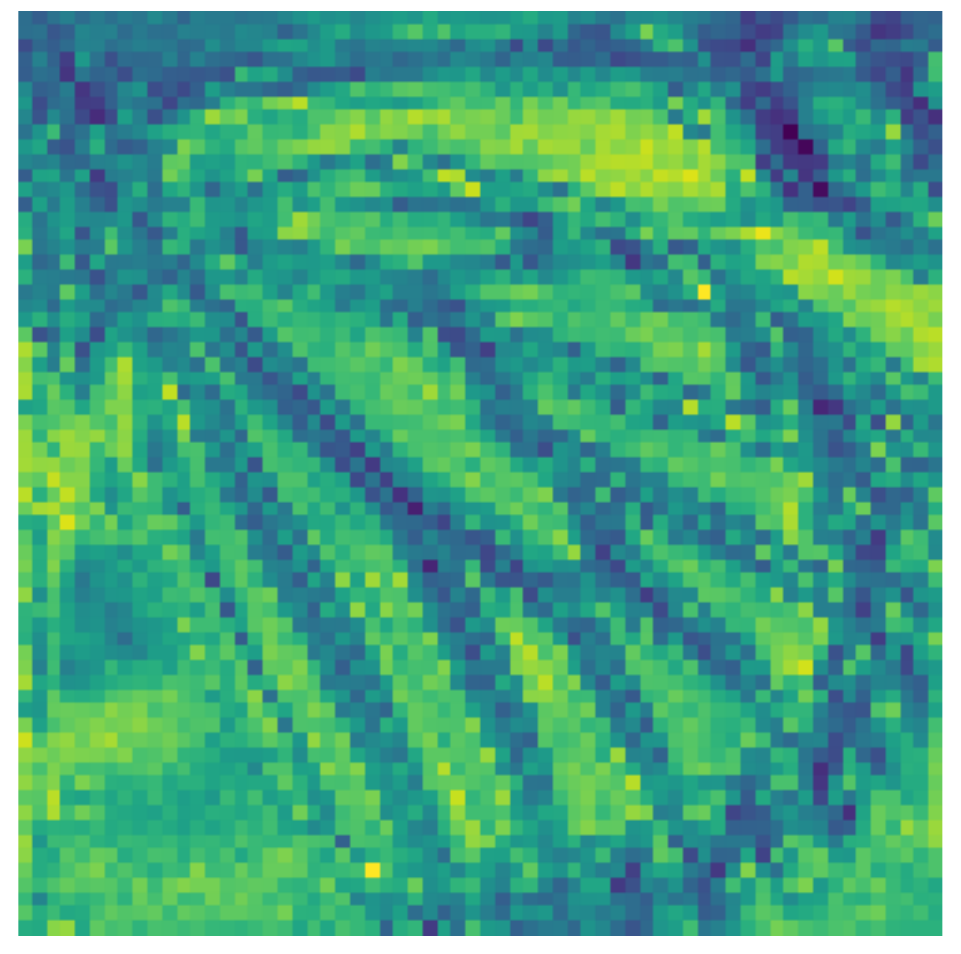}
        \\
        \makebox[0.23\linewidth]{\small (b) w/o SAFM}
        \makebox[0.23\linewidth]{\small (c) w/o FM \& MR \& FA}
        \makebox[0.23\linewidth]{\small (d) w/o FM \& MR}
        \makebox[0.23\linewidth]{\small (e) w/o FM \& FA}
        \\
        \includegraphics[width=0.23\linewidth]{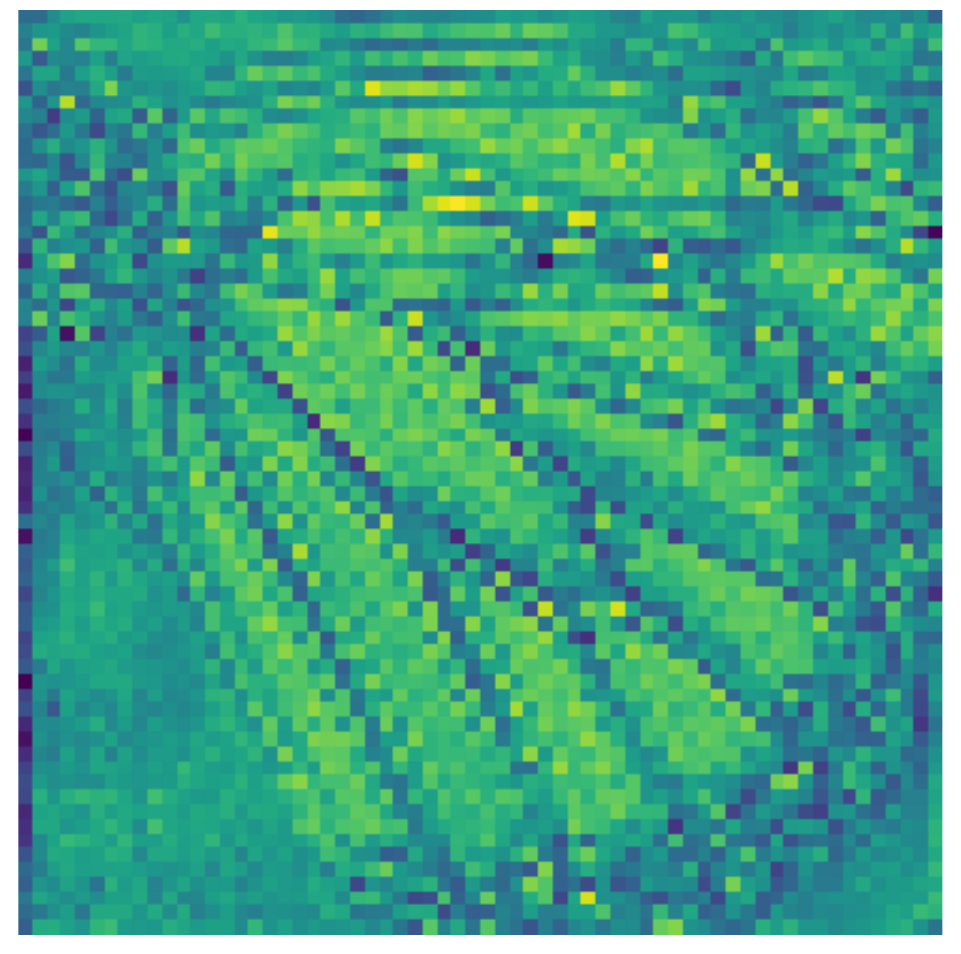}
        \includegraphics[width=0.23\linewidth]{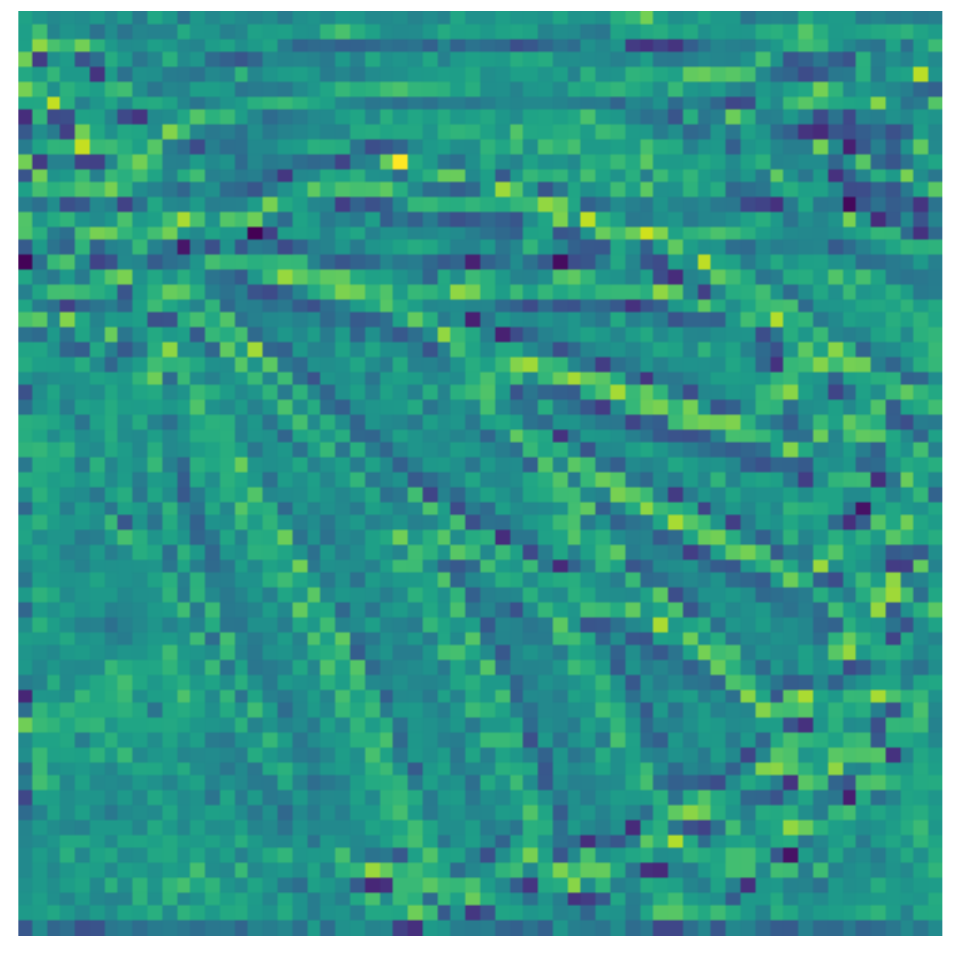}
        \includegraphics[width=0.23\linewidth]{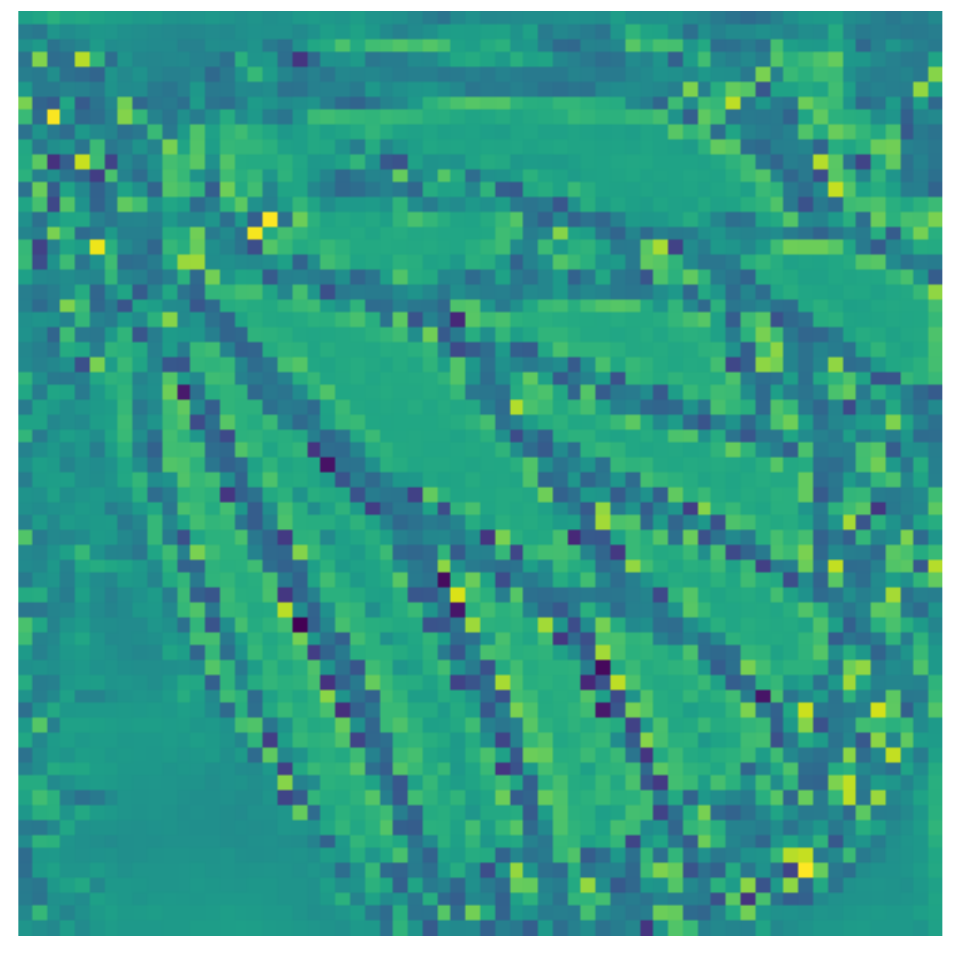}
        \includegraphics[width=0.23\linewidth]{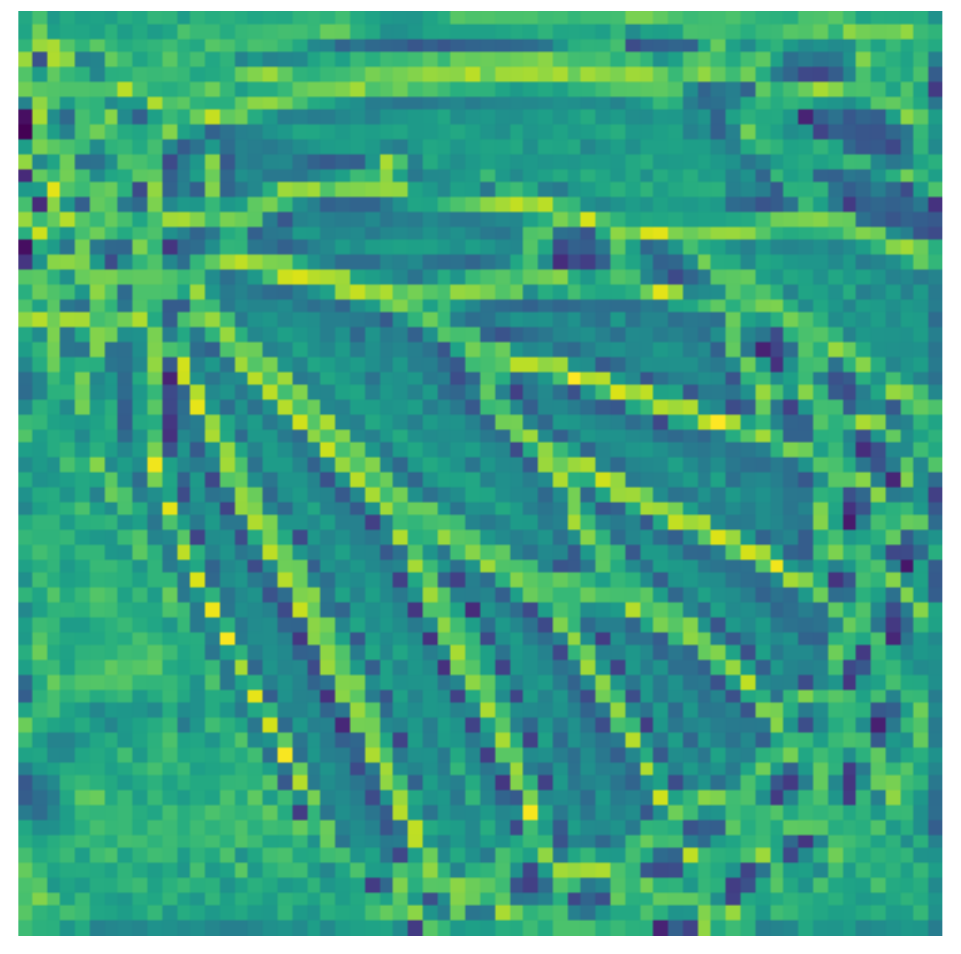}
        \\
        \makebox[0.23\linewidth]{\small (f) w/o FA}
        \makebox[0.23\linewidth]{\small (g) w/o MR}
        \makebox[0.23\linewidth]{\small (h) w/o FM}
        \makebox[0.23\linewidth]{\small (i) w/ SAFM}
	\end{minipage}
    \hspace{-10mm}
    \begin{minipage}{0.03\textwidth} \centering
        \includegraphics[width=2.3\linewidth,height=10\linewidth]{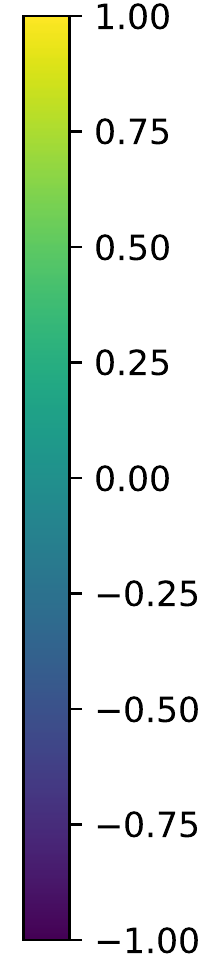}
    \end{minipage}
    \vspace{-1mm}
    \caption{Illustration of learned deep features from the SAFM ablation. We average the features before the upsampling layer in the channel dimension and show the corresponding results.
    The proposed SAFM layer includes three components: feature modulation (FM), multi-scale representation (MR), and feature aggregation (FA).
    (h) indicates that the model pays less attention to the high-frequency regions without the feature modulation.
    (g) shows that the model fails to capture long-range information without the multi-scale representation.
    (f) illustrates the necessity of aggregating multi-scale features.
    The comparison of (i) with (b)-(h) suggests that the proposed method with the SAFM block yields a finer feature representation with clearer structures that pays more attention to high-frequency details.}
    \label{fig:SAFM}
    \vspace{-2mm}
\end{figure*}

\noindent
\textbf{Effectiveness of the spatially-adaptive feature modulation.}
As described in the main paper, the proposed spatially-adaptive feature modulation layer consists of three components: feature modulation (FM), multi-scale representation (MR), and feature aggregation (FA).
To intuitively illustrate what the SAFM block learns, we show some learned features in Figure~\ref{fig:SAFM}, where the corresponding features are extracted before the upsampling layer.
%
%
Figure~\ref{fig:SAFM} demonstrates that the deep features learned with SAFM layers contain much richer feature information and attend to more high-frequency details, facilitating the reconstruction of high-quality images.

\begin{figure*}[t] \centering
    \newcommand{\hwidth}{1pt}
    \includegraphics[width=0.31\textwidth, height=0.20\textwidth]{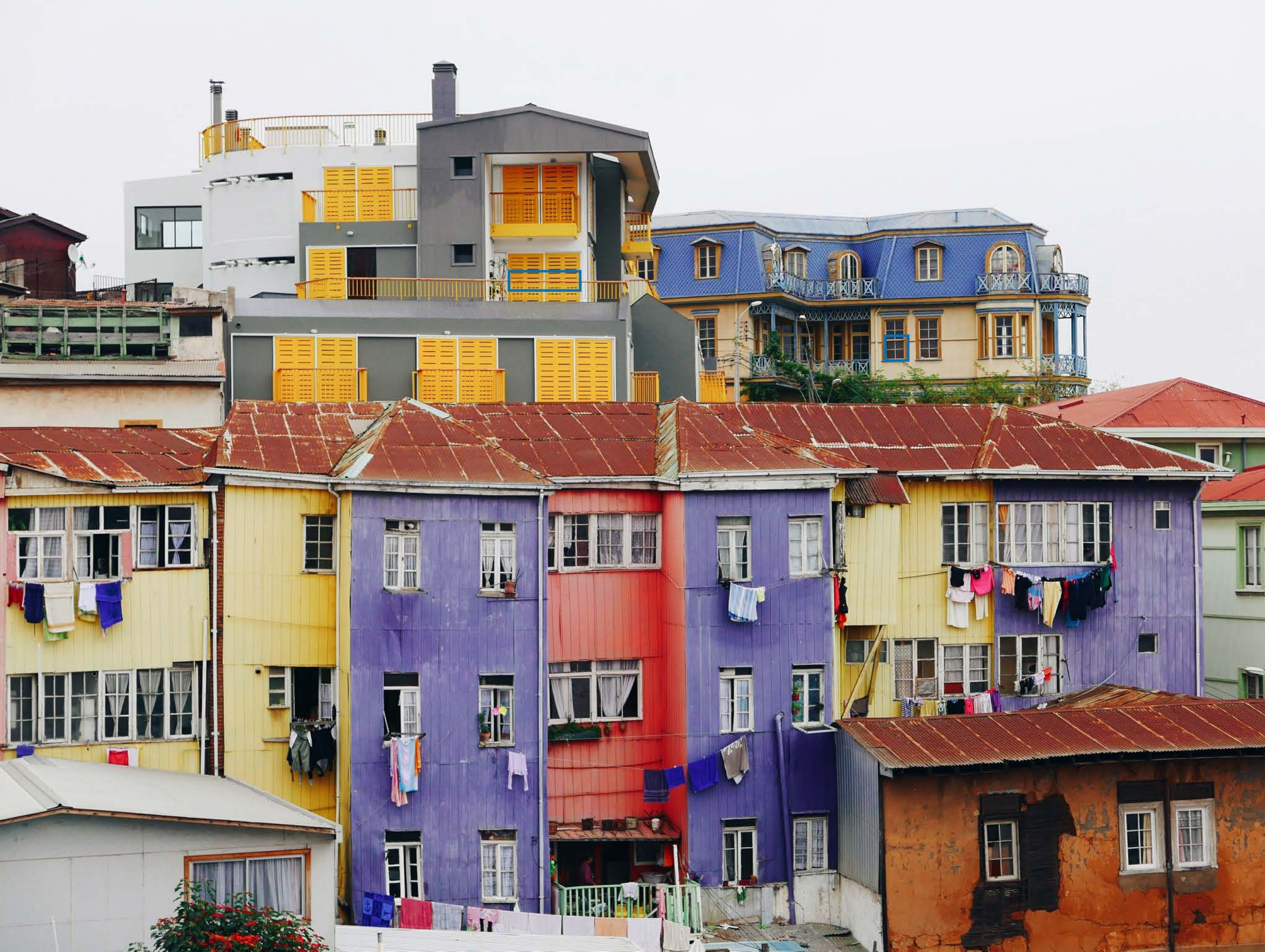}
    \includegraphics[width=0.31\textwidth, height=0.20\textwidth]{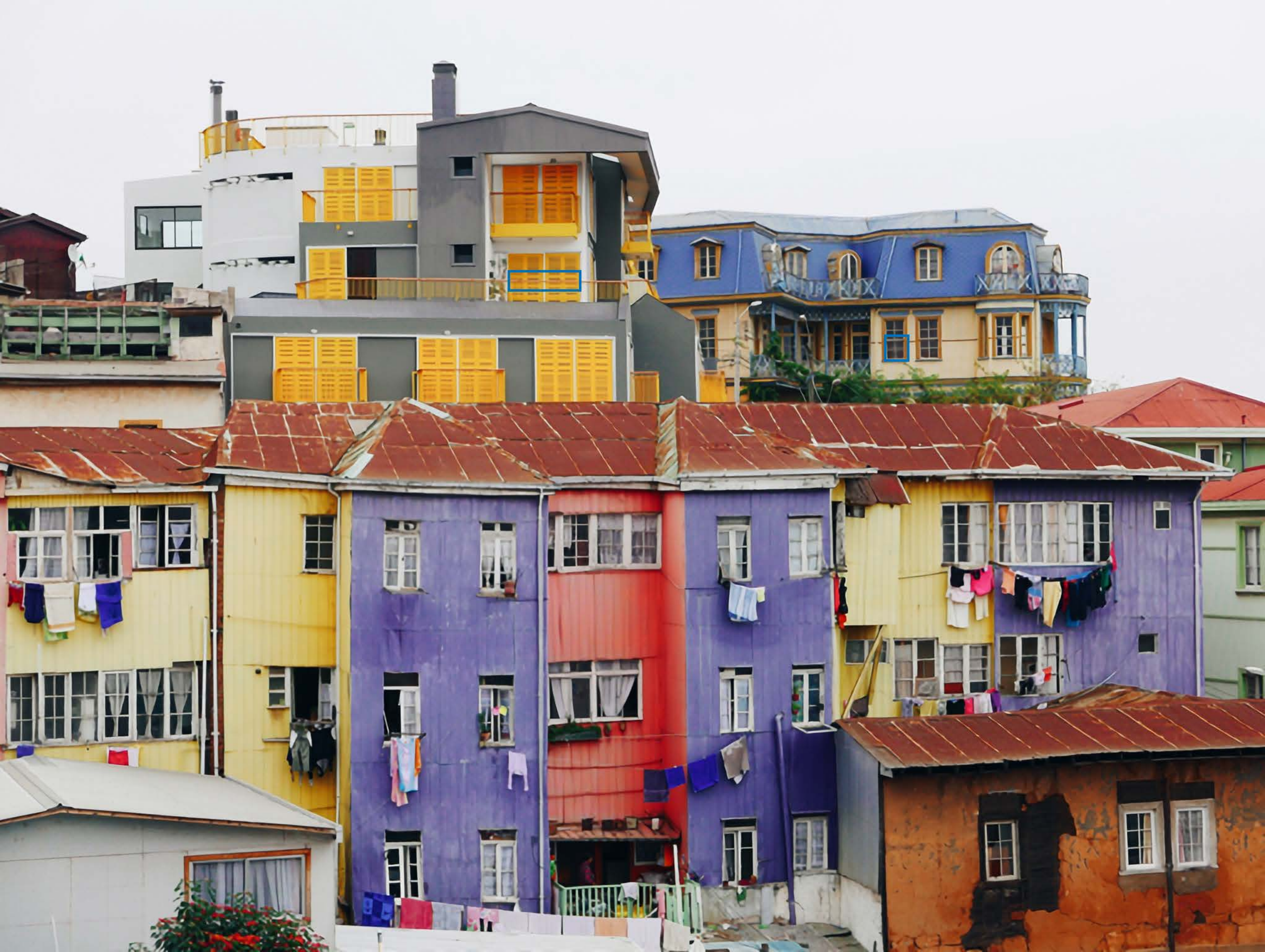}
    \includegraphics[width=0.31\textwidth, height=0.20\textwidth]{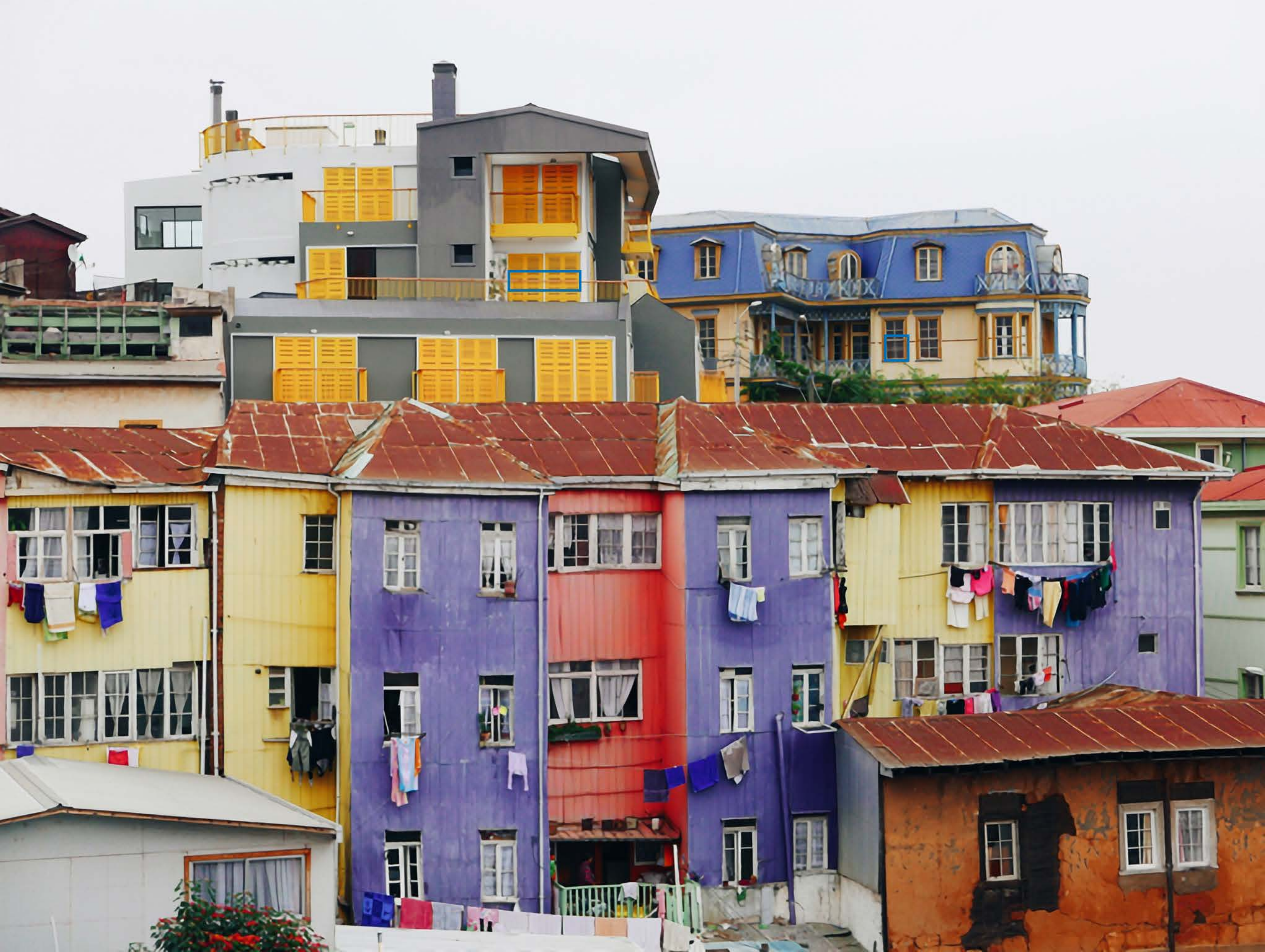}
    \\[2pt]
    \makebox[0.31\textwidth]{%
        \includegraphics[width=0.15\textwidth,height=0.08\textwidth]{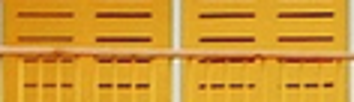}\hspace{\hwidth}
        \includegraphics[width=0.15\textwidth,height=0.08\textwidth]{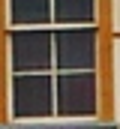}}
    \makebox[0.31\textwidth]{%
        \includegraphics[width=0.15\textwidth,height=0.08\textwidth]{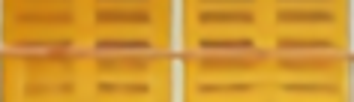}\hspace{\hwidth}
        \includegraphics[width=0.15\textwidth,height=0.08\textwidth]{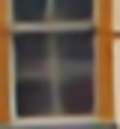}}
    \makebox[0.31\textwidth]{%
        \includegraphics[width=0.15\textwidth,height=0.08\textwidth]{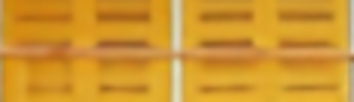}\hspace{\hwidth}
        \includegraphics[width=0.15\textwidth,height=0.08\textwidth]{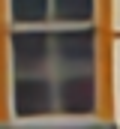}}
    \\
    \includegraphics[width=0.31\textwidth, height=0.20\textwidth]{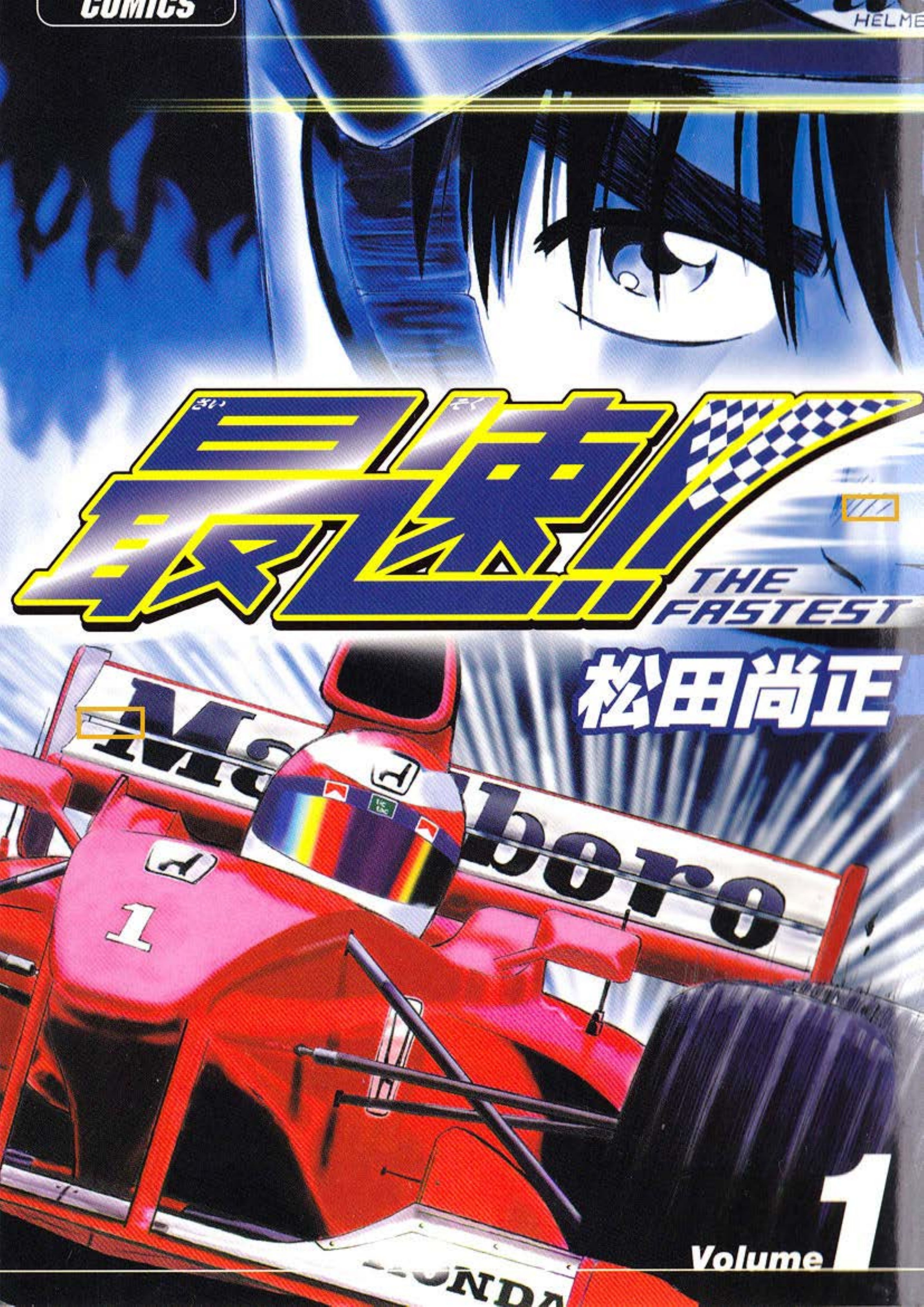}
    \includegraphics[width=0.31\textwidth, height=0.20\textwidth]{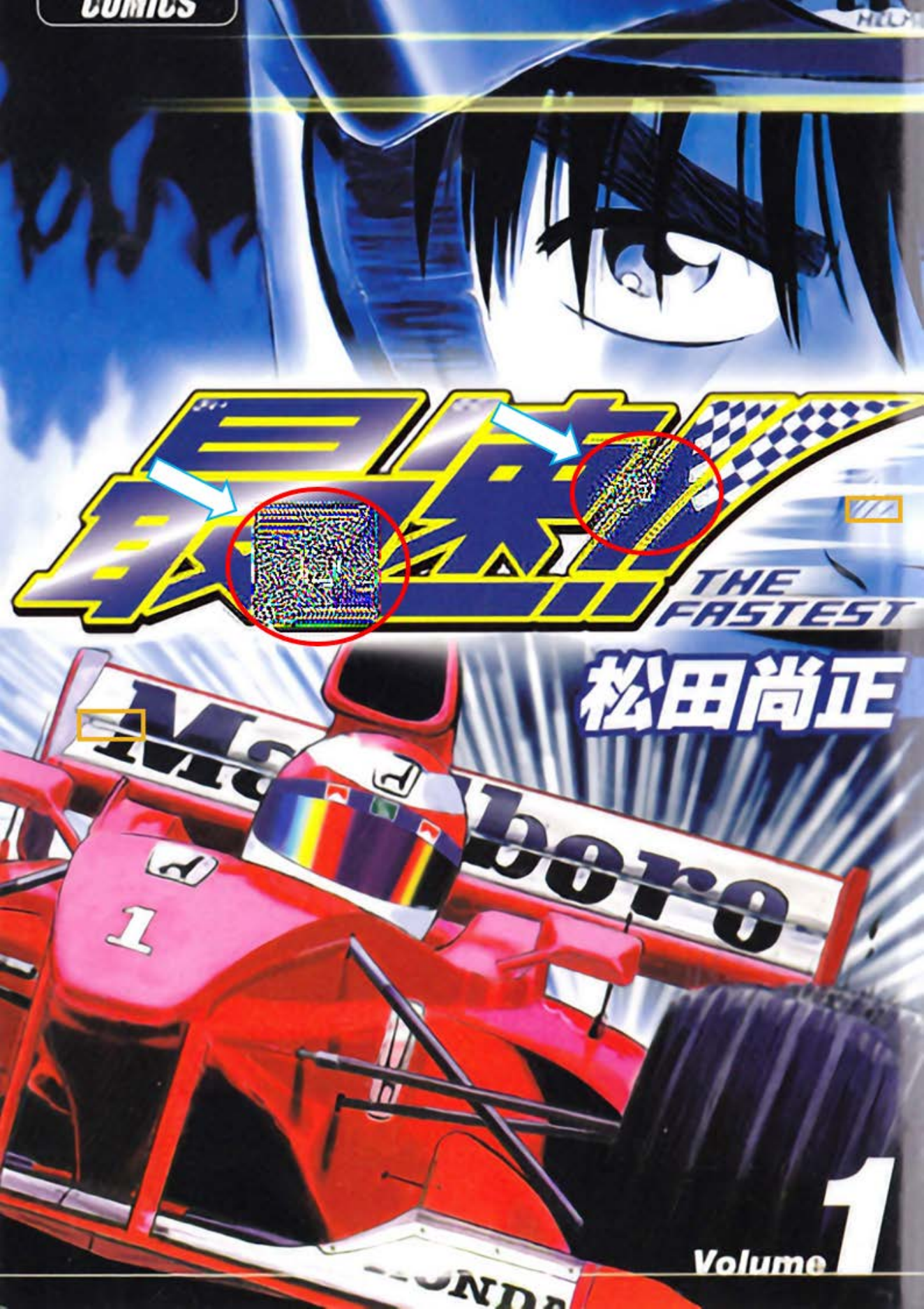}
    \includegraphics[width=0.31\textwidth, height=0.20\textwidth]{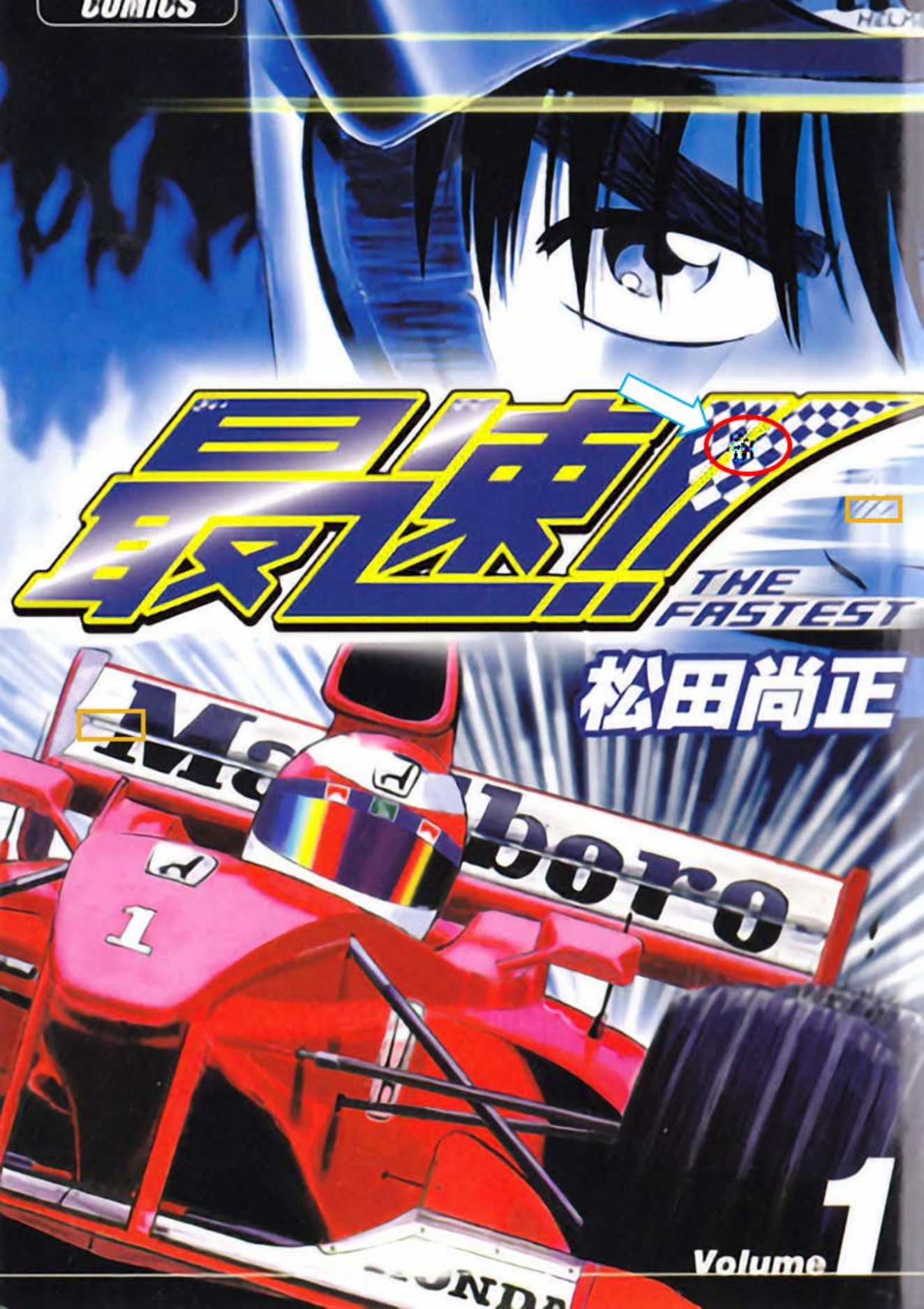}
    \\[2pt]
    \makebox[0.31\textwidth]{%
        \includegraphics[width=0.15\textwidth,height=0.08\textwidth]{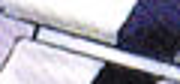}\hspace{\hwidth}
        \includegraphics[width=0.15\textwidth,height=0.08\textwidth]{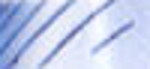}}
    \makebox[0.31\textwidth]{%
        \includegraphics[width=0.15\textwidth,height=0.08\textwidth]{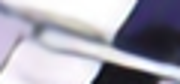}\hspace{\hwidth}
        \includegraphics[width=0.15\textwidth,height=0.08\textwidth]{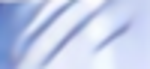}}
    \makebox[0.31\textwidth]{%
        \includegraphics[width=0.15\textwidth,height=0.08\textwidth]{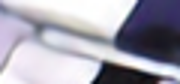}\hspace{\hwidth}
        \includegraphics[width=0.15\textwidth,height=0.08\textwidth]{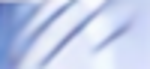}}
    \\
    \makebox[0.31\textwidth]{\small (a) GT}
    \makebox[0.31\textwidth]{\small (b) BatchNorm~\cite{BN}}
    \makebox[0.31\textwidth]{\small (c) Frozen BatchNorm~\cite{BN}}

    \includegraphics[width=0.31\textwidth, height=0.20\textwidth]{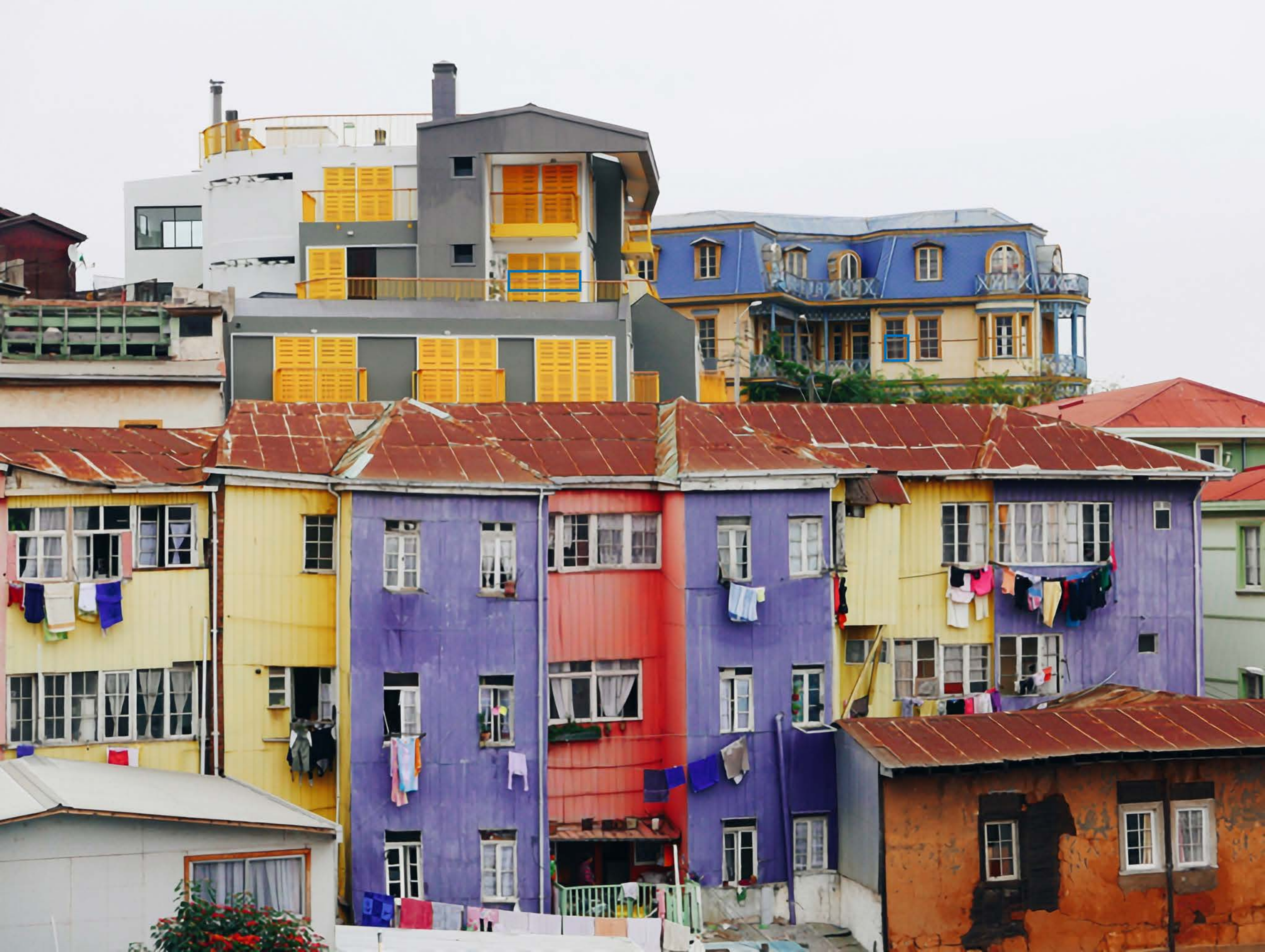}
    \includegraphics[width=0.31\textwidth, height=0.20\textwidth]{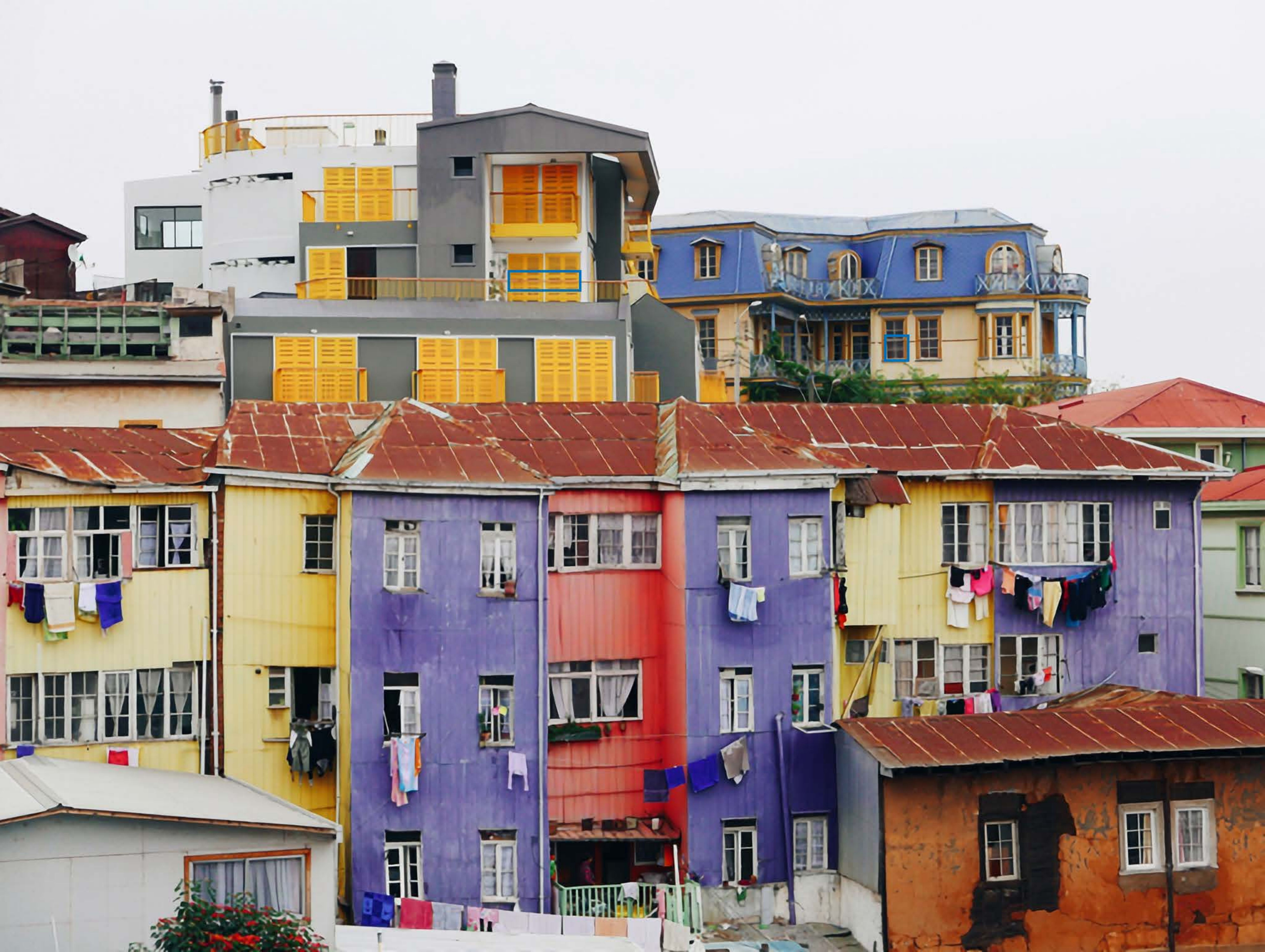}
    \includegraphics[width=0.31\textwidth, height=0.20\textwidth]{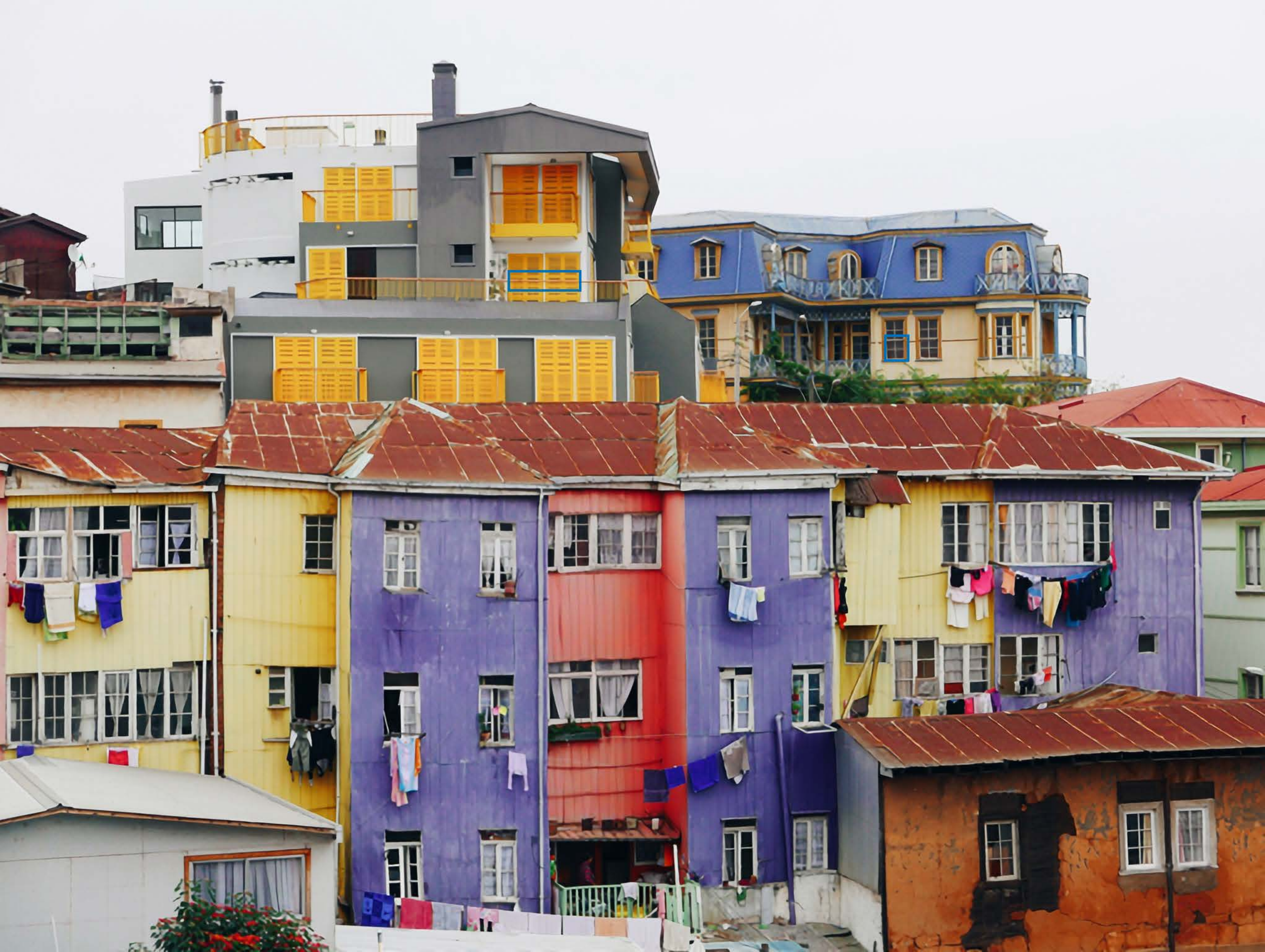}
    \\[2pt]
    \makebox[0.31\textwidth]{%
        \includegraphics[width=0.15\textwidth,height=0.08\textwidth]{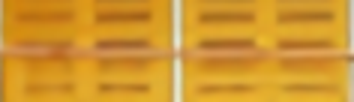}\hspace{\hwidth}
        \includegraphics[width=0.15\textwidth,height=0.08\textwidth]{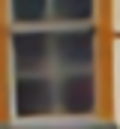}}
    \makebox[0.31\textwidth]{%
        \includegraphics[width=0.15\textwidth,height=0.08\textwidth]{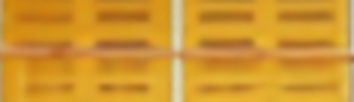}\hspace{\hwidth}
        \includegraphics[width=0.15\textwidth,height=0.08\textwidth]{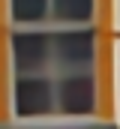}}
    \makebox[0.31\textwidth]{%
        \includegraphics[width=0.15\textwidth,height=0.08\textwidth]{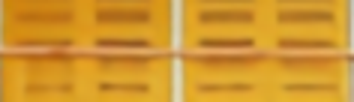}\hspace{\hwidth}
        \includegraphics[width=0.15\textwidth,height=0.08\textwidth]{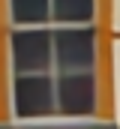}}
    \\
    \includegraphics[width=0.31\textwidth, height=0.20\textwidth]{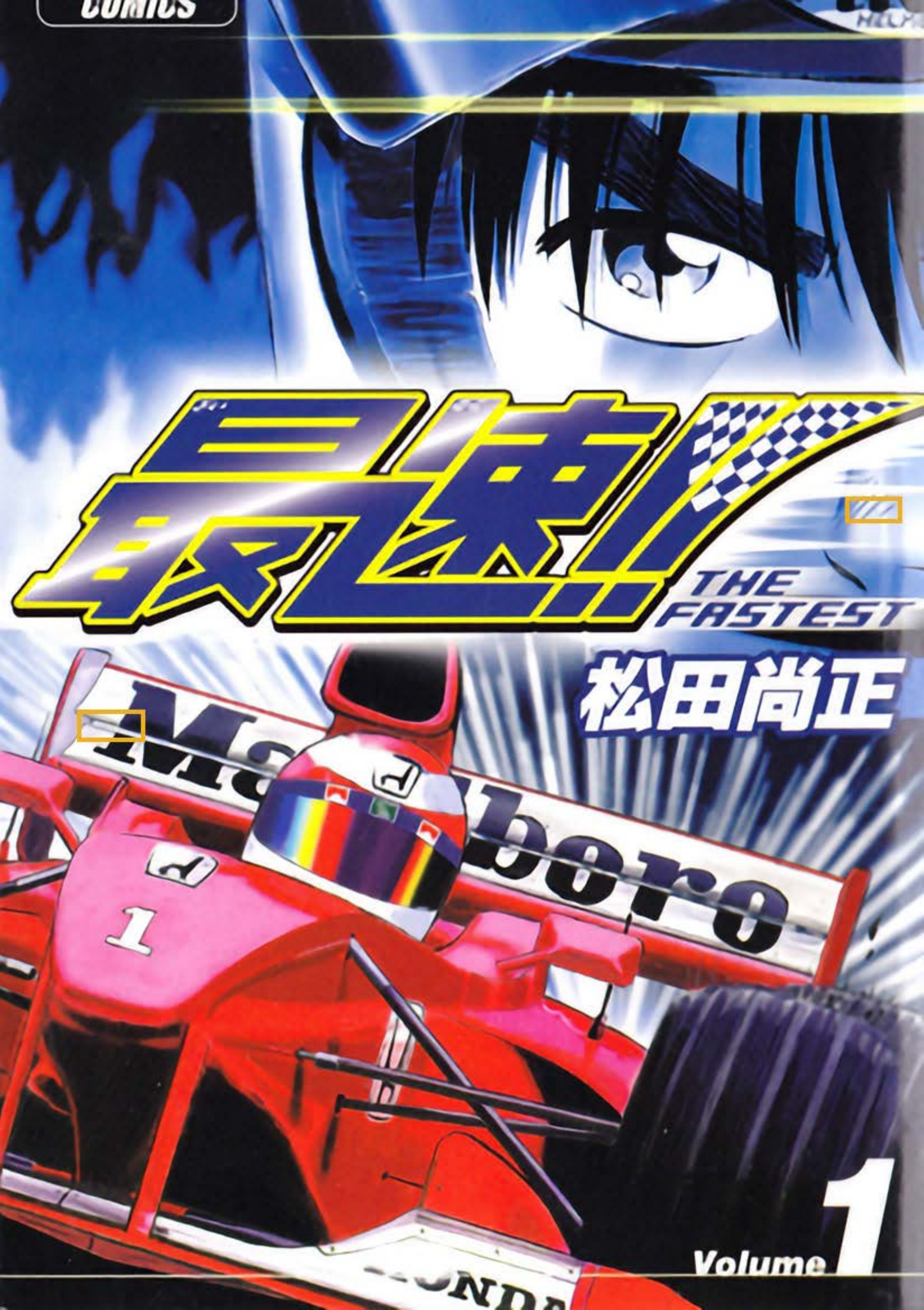}
    \includegraphics[width=0.31\textwidth, height=0.20\textwidth]{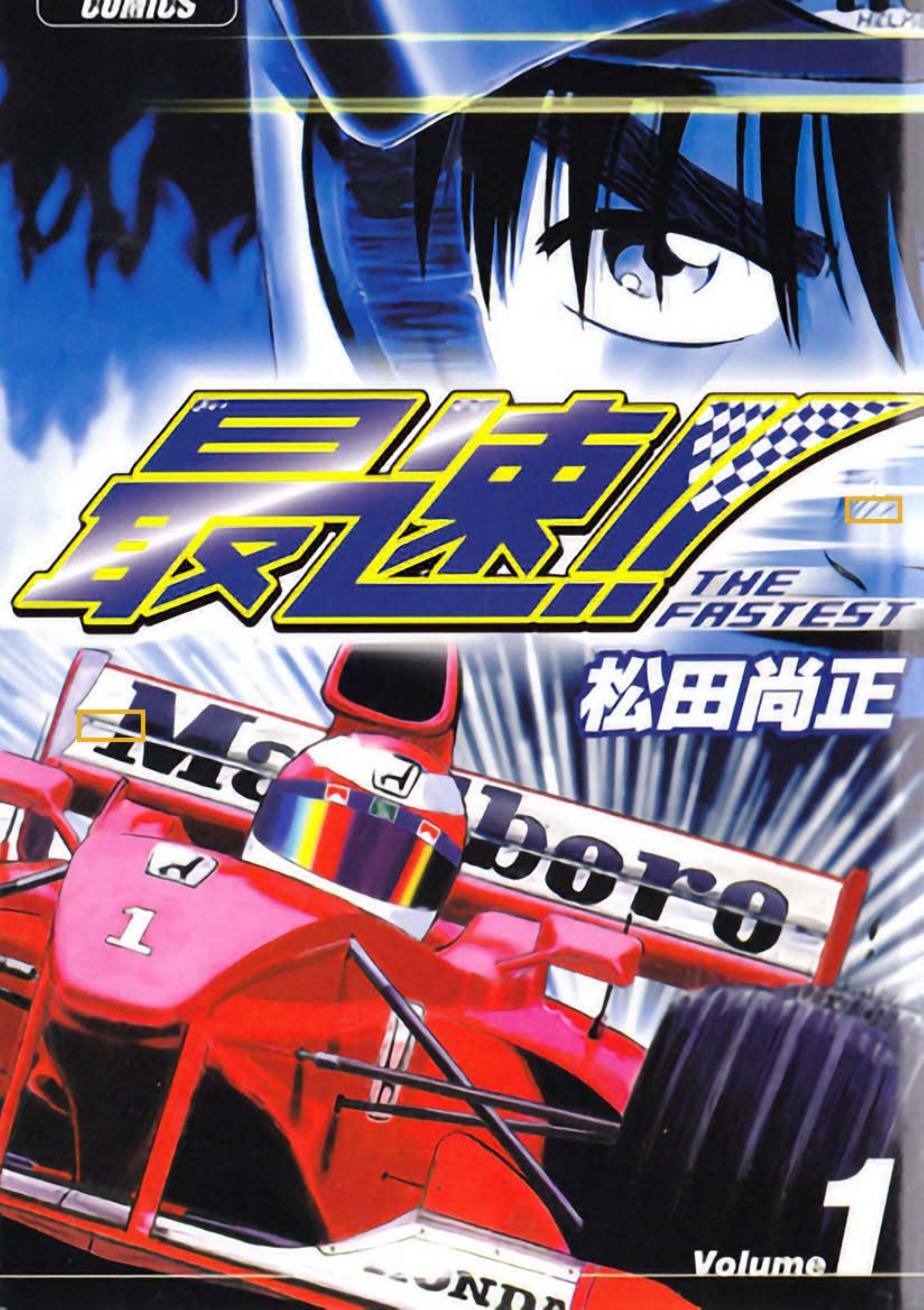}
    \includegraphics[width=0.31\textwidth, height=0.20\textwidth]{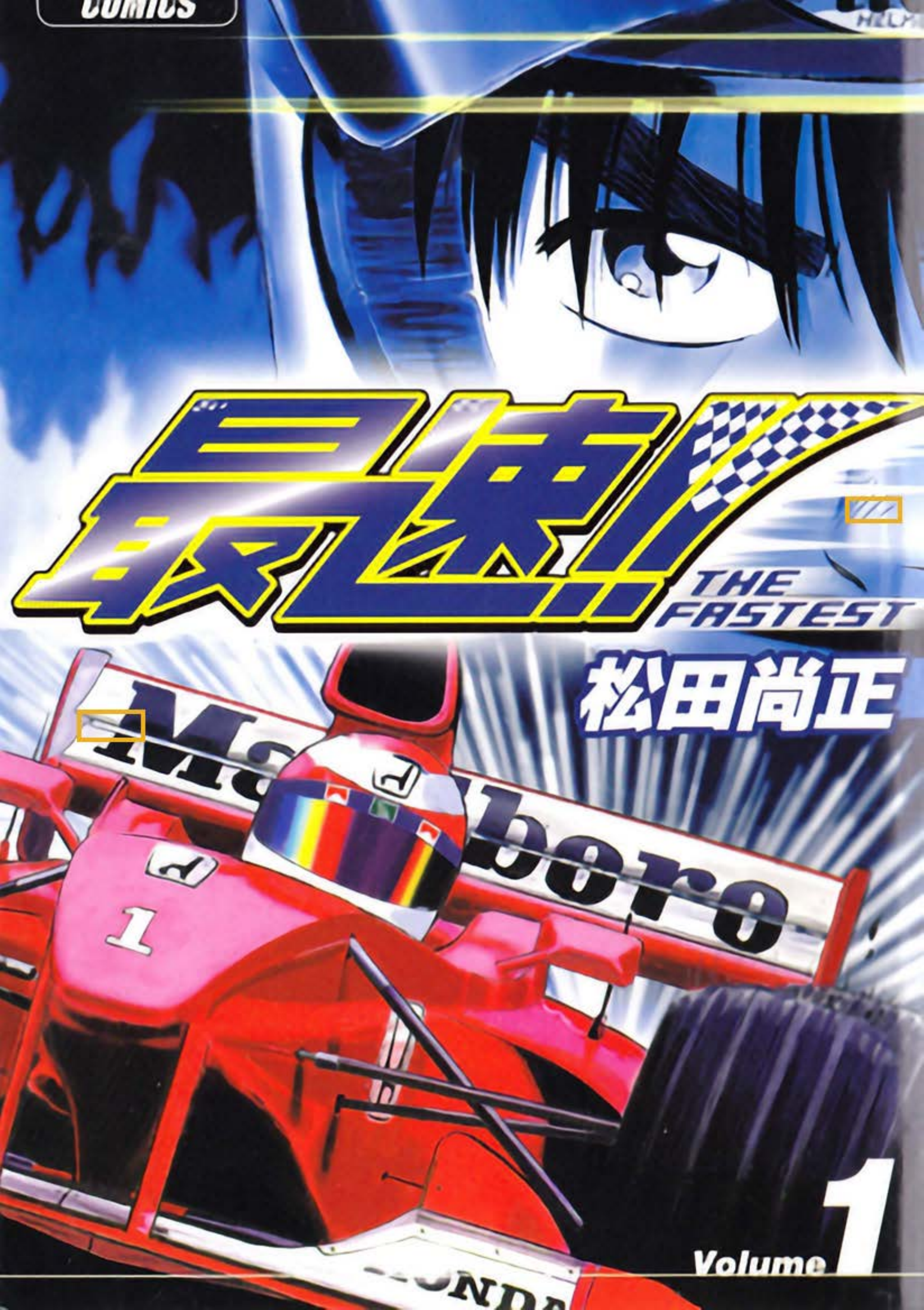}
    \\[2pt]
    \makebox[0.31\textwidth]{%
        \includegraphics[width=0.15\textwidth,height=0.08\textwidth]{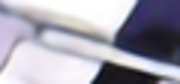}\hspace{\hwidth}
        \includegraphics[width=0.15\textwidth,height=0.08\textwidth]{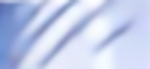}}
    \makebox[0.31\textwidth]{%
        \includegraphics[width=0.15\textwidth,height=0.08\textwidth]{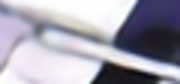}\hspace{\hwidth}
        \includegraphics[width=0.15\textwidth,height=0.08\textwidth]{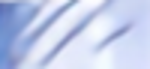}}
    \makebox[0.31\textwidth]{%
        \includegraphics[width=0.15\textwidth,height=0.08\textwidth]{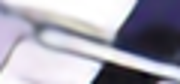}\hspace{\hwidth}
        \includegraphics[width=0.15\textwidth,height=0.08\textwidth]{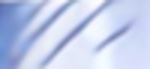}}
    \\
    \makebox[0.31\textwidth]{\small (d) $\text{L}_2$ normalization }
    \makebox[0.31\textwidth]{\small (e) w/o LayerNorm~\cite{LN}}
    \makebox[0.31\textwidth]{\small (f) w/ LayerNorm~\cite{LN}}
    \\
    \caption{Visual results of different normalization methods. The proposed model with LayerNorm layers reconstructs better images.}
    \label{fig:Norm}
\end{figure*}

\noindent
\textbf{Effectiveness of the LayerNorm layer.}
We show the visual results of different normalizations in Figure~\ref{fig:Norm}.
As stated in the main paper, we obtained the results for the Frozen BatchNorm~\cite{BN} and without the LayerNorm~\cite{LN} before the training collapse.
The model with BatchNorm layers generates images with unpleasant artifacts because it involves the estimated mean and variance of the entire training dataset during testing.
The artifacts can be alleviated when we fix these estimates, as shown in Figure~\ref{fig:Norm}(c).
Figure~\ref{fig:Norm}(d) and (f) show that applying normalization in the channel dimension can avoid the occurrence of artifacts.
Compared to the $\text{L}_2$ normalization, the model with LayerNorm layers produces more precise results.
We, therefore, introduce LayerNorm layers for stable training and well convergence.
The reasons behind the LayerNorm remain to be further investigated.

\section{Comparison with the challenge winners}
\label{sec: challenge}
We further compare our method with solutions of the challenge champion, i.e., RFDN (winner of AIM 2020 Efficient Super-Resolution Challenge~\cite{aim_efficientsr}) and RLFN (winner of NTIRE 2022 Efficient Super-Resolution Challenge~\cite{ntire22efficientsr}).
Table~\ref{tab:runtime} demonstrates that our approach obtains a noticeable improvement in all measures except the running time.
Table~\ref{tab:meta_time} shows that our slower running time is mainly due to the use of LayerNorm~\cite{LN}, which requires the mean and standard deviation of the input features in the inference phase.
Without LayerNorm, the runtime improves to 8.35ms, which is very close to the speed of RLFN.
As shown in Section~\ref{sec: ablation}, however, the importance of LayerNorm prevents us from removing this module directly. We will explore feasible alternatives in our further work.
Moreover, we count the inference time of different commonly used tensor operations with the \href{https://pytorch.org/tutorials/recipes/recipes/profiler_recipe.html}{torch.profiler} function to explicitly evaluate factors affecting the speed of our method.
Table~\ref{tab:meta_time} presents that element-wise addition, element-wise product, and channel splitting are rather time-consuming.
Since we use feature modulation and residual learning in each building block to learning high-frequency details, which also deteriorates the inference speed.

\begin{table*}
	\caption{Effect of scales in the SAFM.
    We evaluate the effect of features at different scales in the SAFM on the ×4 DIV2K validation set. The results show that removing any scale information affects the reconstruction performance.
    }
	\vspace{-2mm}
	\centering
	\tiny
    \resizebox{0.96\linewidth}{!}{
	\begin{tabular}{|l|c|c|c|c|}
		\hline
	  ~Variants~  & SAFMN & ~w/o Scale 8~  & w/o Scale 8\&4  & w/o Scale 8\&4\&2\\
        \hline
        DIV2K\_val    &30.43/0.8372  &30.39/0.8362 &30.37/0.8357 &30.34/0.8350 \\
		\hline
	\end{tabular}}
	\label{tab:Scale}
\end{table*}

\begin{table*}
    \caption{Efficiency comparison with the challenge winners on $\times 4$ SR. \#GPU Mem. and \#Avg. Time denote the maximum GPU memory consumption and the average running time of the inference phase, respectively. \#FLOPs, \#Acts and \#Avg. Time are computed on an LR image with a resolution of $320\times180$ pixels. Our SAFMN obtains comparable performance and a better trade-off between reconstruction performance and model complexity.}
    \vspace{-2mm}
    \centering
    \small
    \resizebox{0.96\linewidth}{!}{
    \begin{tabular}{|l|c|c|c|c|c|c|}
        \hline
        Methods     &\#Params [K]   &\#FLOPs [G]   &\#Acts [M]  &\#GPU Mem. [M]   &\#Avg.Time [ms] &B100 [PSNR/SSIM]\\
         \hline
        RFDN~\cite{rfdn}      & 433.45  & 23.82  & 98.46    & 176.75  & 7.23        &27.60/0.7368\\
        RLFN~\cite{rlfn}      & 543.74  & 29.88  & 111.17   & 145.69  & 7.35        &27.60/0.7364 \\
        SAFMN                 & 239.52  & 13.56  & 76.70    & 65.26   & 10.71       &27.58/0.7359\\
        \hline
    \end{tabular}}
    \label{tab:runtime}
\end{table*}

\begin{table}
	\caption{GPU runtime of commonly used meta-operations on $\times4$ SR.
    we count the time with the \href{https://pytorch.org/tutorials/recipes/recipes/profiler_recipe.html}{torch.profiler} function.
    }
	\vspace{-2mm}
	\centering
	\small
    \resizebox{0.96\linewidth}{!}{
    	\begin{tabular}{|c|c|c|c|c|c|c|c|c|}
    		\hline
    	  Meta-operations & LayerNorm & Channel Splitting  &Adaptive MaxPool  & Nearest Interpolation &Element-wise Product & Element-wise Addition\\
            \hline
            Runtime [us]    &465  &133  &85 &24 &156 &193\\
    		\hline
    	\end{tabular}}
	\label{tab:meta_time}
\end{table}

\section{Comparison with classic SR models}
\label{sec: classic_sr}
To verify the scalability of SAFMN, we further compare the large version of SAFMN with the state-of-the-art classical SR methods, including EDSR~\cite{EDSR}, RCAN~\cite{RCAN}, SAN~\cite{SAN}, HAN~\cite{HAN}, SwinIR~\cite{SwinIR}.
Table~\ref{tab:classic_sr} shows that our SAFMN shows significant advantages in terms of model efficiency compared to the evaluated CNN-based methods and obtains competitive reconstruction performances, benefiting from its adaptability upon a multi-scale feature representation-based modulation mechanism.

\begin{table*}[t]
	\caption{Comparison with classic SR models.
        \#Params and \#FLOPs are measured under the setting of upscaling SR images to 1280 $\times$ 720 pixels
on all scales.
        The proposed SAFMN achieves comparable performances with significantly less computational and memory costs.}
	\label{tab:classic_sr}
	\vspace{-5mm}
	\begin{center}
		\resizebox{0.96\linewidth}{!}{
			\begin{tabular}{| l | c | c | c | c | c | c | c |c|}
				\hline
				Scale & Methods & \#Params [M] & \#FLOPs [G] & Set5 & Set14 & B100 & Urban100 & Manga109 \\
				\hline
				\multirow{6}*{$\times 2$}  & EDSR~\cite{EDSR} &40.73    &9387   &38.11/0.9602 &33.92/0.9195 &32.32/0.9013 &32.93/0.9351 &39.10/0.9773\\
				~   &RCAN~\cite{RCAN} &15.45    &3530   &38.27/0.9614 &34.12/0.9216 &32.41/0.9027 &33.34/0.9384 &39.44/0.9786\\
                ~   &SAN~\cite{SAN}   &15.86    &3050   &38.31/0.9620 &34.07/0.9213 &32.42/0.9028 &33.10/0.9370 &39.32/0.9792\\
				~   &HAN~\cite{HAN}   &63.61    &14551   &38.27/0.9614 &34.16/0.9217 &32.41/0.9027 &33.35/0.9385 &39.46/0.9785\\
				~   &\textbf{SAFMN}           &\textbf{5.56}    &\textbf{1274}   &38.28/0.9616 &34.14/0.9220 &32.39/0.9024 &33.06/0.9366 &39.56/0.9790\\
                \cdashline{2-9}[1pt/1pt]
                ~   &SwinIR~\cite{SwinIR} &11.75    &2301   &38.42/0.9623  &34.46/0.9250  &32.53/0.9041 &33.81/0.9427 &39.92/0.9797\\

				\hline
				\multirow{6}*{$\times 3$} &EDSR~\cite{EDSR}    &43.68    &4470   &34.65/0.9280 &30.52/0.8462 &29.25/0.8093 &28.80/0.8653 &34.17/0.9476\\
				~  &RCAN~\cite{RCAN}  &15.63    &1586   &34.65/0.9280 &30.52/0.8462 &29.25/0.8093 &28.80/0.8653 &34.17/0.9476\\
                ~  &SAN~\cite{SAN}    &15.90    &1620   &34.75/0.9300 &30.59/0.8476 &29.33/0.8112 &28.93/0.8671 &34.30/0.9494\\
				~  &HAN~\cite{HAN}    &64.35    &6534   &34.75/0.9299 &30.67/0.8483 &29.32/0.8110 &29.10/0.8705 &34.48/0.9500\\
				~  &\textbf{SAFMN}  &\textbf{5.58}    &\textbf{569}   &34.80/0.9301  &30.68/0.8485  &29.34/0.8110  &28.99/0.8679  &34.66/0.9504 \\
                \cdashline{2-9}[1pt/1pt]
                ~  &SwinIR~\cite{SwinIR}  &11.94    &1026   &34.97/0.9318 &30.93/0.8534 &29.46/0.8145 &29.75/0.8826 &35.12/0.9537\\
				
				\hline
				\multirow{6}*{$\times 4$}  &EDSR~\cite{EDSR}    &43.90    &2895   &32.46/0.8968 &28.80/0.7876  &27.71/0.7420  &26.64/0.8033  &31.02/0.9148\\
				~  &RCAN~\cite{RCAN}   &15.59    &918   &32.63/0.9002 &28.87/0.7889 &27.77/0.7436 &26.82/0.8087 &31.22/0.9173\\
                ~  &SAN~\cite{SAN}     &15.86    &937   &32.64/0.9003 &28.92/0.7888 &27.78/0.7436 &26.79/0.8068 &31.18/0.9169\\
				~  &HAN~\cite{HAN}     &64.20   &3776   &32.64/0.9002 &28.90/0.7890 &27.80/0.7442 &26.85/0.8094 &31.42/0.9177\\
				~  &\textbf{SAFMN}              &\textbf{5.60}    &\textbf{321}   &32.65/0.9005   &28.96/0.7898   &27.82/0.7440 &26.81/0.8058 &31.59/0.9192\\
				\cdashline{2-9}[1pt/1pt]
                ~  &SwinIR~\cite{SwinIR} &11.90    &834   &32.92/0.9044 &29.09/0.7950 &27.92/0.7489 &27.45/0.8254 &32.03/0.9260\\
                \hline
		\end{tabular}}
	\end{center}
	\vspace{-5mm}
\end{table*}

\section{Some notes on the Urban100 dataset}
\label{sec: urban100_notes}
As shown in Table~\ref{tab:urban100}, the proposed SAFMN obtains a weak PSNR performance on the Urban100 dataset compared to other state-of-the-art methods, e.g., IMDN~\cite{IMDN} and LAPAR-A~\cite{LAPAR}.
The slight local luminance differences are responsible for these results.
Since PSNR measures pixel-level distances rather than overall structure, slight differences in the luminance channel can lead to significant differences in PSNR.
Furthermore, we visually compare images with a significant PSNR gap between our SAFMN and IMDN and observe no detectable changes in perceptual quality.
Thus, we reevaluate these results using two commonly-used perceptual metrics: NIQE and LPIPS.
Table~\ref{tab:urban100} lists the quantitative results, and the proposed method achieves comparable performance to IMDN and LAPAR-A in terms of NIQE and LPIPS.

\begin{table}[t]
	\caption{Quantitative comparison results on the Urban100 dataset. Our proposed method performs weakly in PSNR/SSIM but is comparable to IMDN and LAPAR in perceptual metrics, including NIQE and LPIPS.}
    \vspace{-2mm}
    \small
	\centering
    \resizebox{0.96\linewidth}{!}{
    \begin{tabular}{|c |l |c |c |c | c | c | c | }
        \hline
        Scale &Methods &\#Params [K] &\#FLOPs [G] &PSNR &SSIM &NIQE &LPIPS \\
        \hline
        \multirow{3}*{$\times 2$} &IMDN~\cite{IMDN} & 694 & 161 & 32.17 & 0.9283 & 4.59  & 0.1132   \\
        ~ & LAPAR-A~\cite{LAPAR}  & 548 & 171 & 32.17 &0.9250 & 4.55  & 0.1129 \\
        ~ & SAFMN                 & 228 & 52 & 31.84 & 0.9256 & 4.60  & 0.1138  \\
        \hline
        \multirow{3}*{$\times 3$} &IMDN~\cite{IMDN} & 703 & 72 & 28.17 & 0.8519 & 5.21  & 0.2136  \\
        ~ & LAPAR-A~\cite{LAPAR}   & 594 & 114 & 28.15 & 0.8523 & 5.21  & 0.2163 \\
        ~ & SAFMN                  & 233 & 23 & 27.95 & 0.8474 & 5.28  & 0.2134  \\
         \hline
        \multirow{3}*{$\times 4$} &IMDN~\cite{IMDN} & 715 & 41 & 26.04 & 0.7838 & 5.69  & 0.2879  \\
        ~ &LAPAR-A~\cite{LAPAR}  & 659 & 94 & 26.14 & 0.7871 & 5.63  & 0.2868  \\
      ~ &SAFMN                  & 240 & 14 & 25.97 & 0.7809 & 5.79  & 0.2881 \\
        \hline
    \end{tabular}}
    \label{tab:urban100}
    \vspace{-5mm}
\end{table}

\section{More visual results}
\label{sec: visual}
\begin{figure*}
	\footnotesize
	\begin{center}
		\begin{tabular}{c c c c c c c c}
			\multicolumn{3}{c}{\multirow{5}*[75pt]{\includegraphics[width=0.35\linewidth, height=0.355\linewidth]{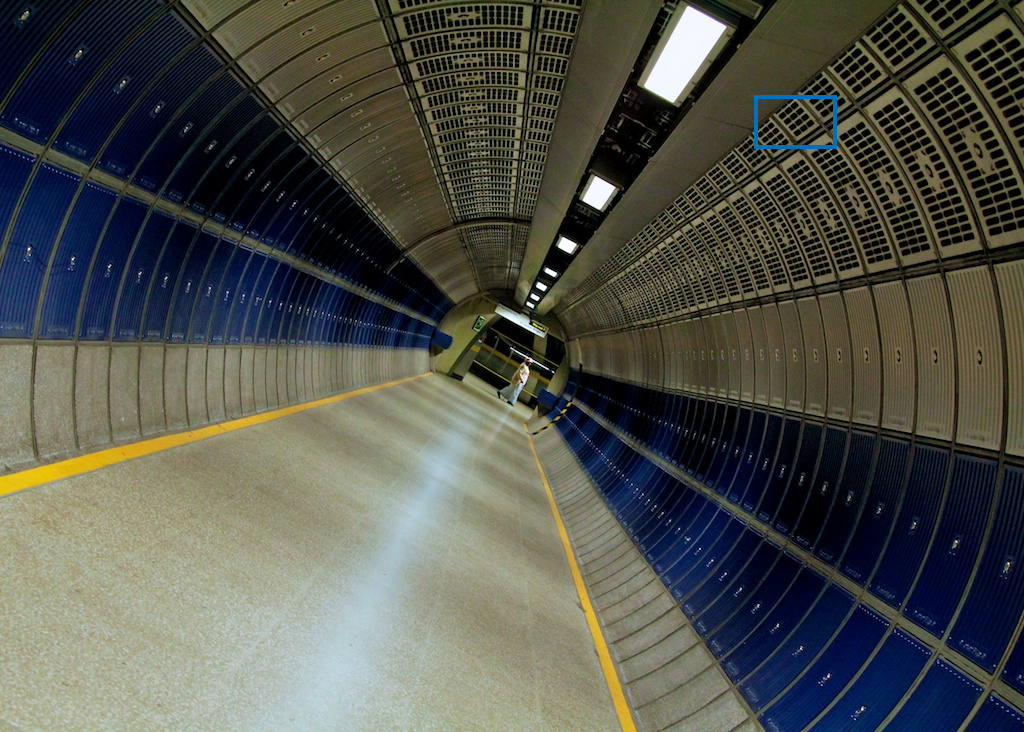}}}&\hspace{-3.0mm}
			\includegraphics[width=0.15\linewidth, height = 0.165\linewidth]{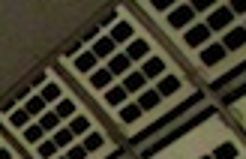} &\hspace{-3.0mm}
			\includegraphics[width=0.15\linewidth, height = 0.165\linewidth]{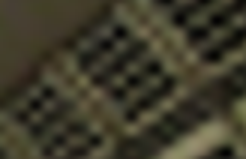}  &\hspace{-3.0mm}
			\includegraphics[width=0.15\linewidth, height = 0.165\linewidth]{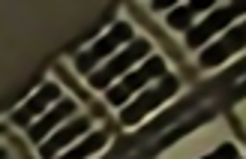}  &\hspace{-3.0mm}
			\includegraphics[width=0.15\linewidth, height = 0.165\linewidth]{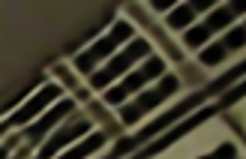} \\
			\multicolumn{3}{c}{~} &\hspace{-3.0mm}  (a) HR patch &\hspace{-3.0mm}  (b) Bicubic &\hspace{-3.0mm}  (c) VDSR~\cite{VDSR}  &\hspace{-3.0mm}  (d) ShuffleMixer~\cite{ShuffleMixer}\\
			
			\multicolumn{3}{c}{~} & \hspace{-3.0mm}
			\includegraphics[width=0.15\linewidth, height = 0.165\linewidth]{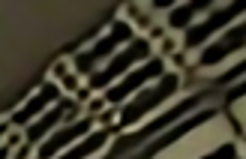} & \hspace{-3.0mm}
			\includegraphics[width=0.15\linewidth, height = 0.165\linewidth]{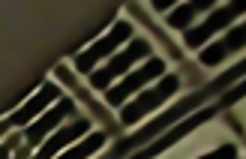} & \hspace{-3.0mm}
			\includegraphics[width=0.15\linewidth, height = 0.165\linewidth]{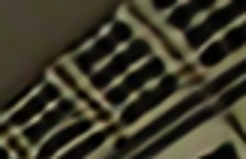} & \hspace{-3.0mm}
			\includegraphics[width=0.15\linewidth, height = 0.165\linewidth]{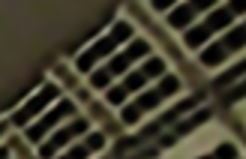} \\
			\multicolumn{3}{c}{\hspace{-3.0mm} img078 from Urban100} &\hspace{-3.0mm}  (e) LapSRN~\cite{LapSRN} &  \hspace{-3.0mm} (f) CARN~\cite{CARN}  &\hspace{-3.0mm}  (g) IMDN~\cite{IMDN}  & \hspace{-3.0mm}(h) SAFMN \\
			
			\\
            \multicolumn{3}{c}{\multirow{5}*[75pt]{\includegraphics[width=0.35\linewidth, height=0.355\linewidth]{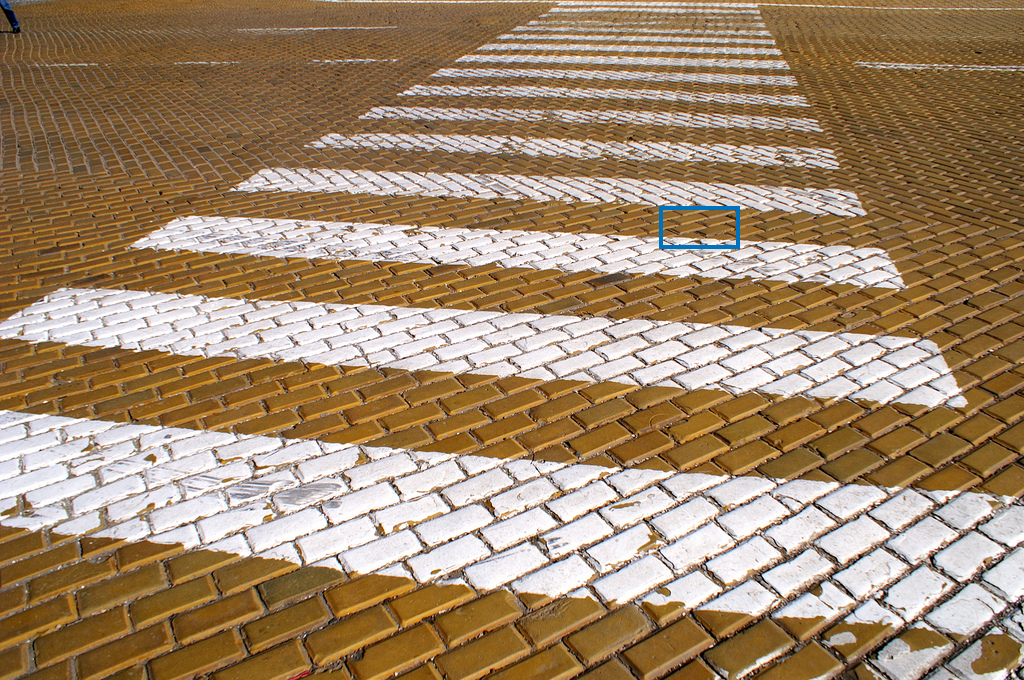}}}&\hspace{-3.0mm}
			\includegraphics[width=0.15\linewidth, height = 0.165\linewidth]{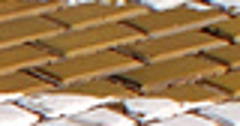} &\hspace{-3.0mm}
			\includegraphics[width=0.15\linewidth, height = 0.165\linewidth]{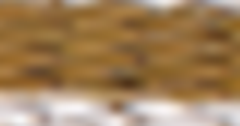}  &\hspace{-3.0mm}
			\includegraphics[width=0.15\linewidth, height = 0.165\linewidth]{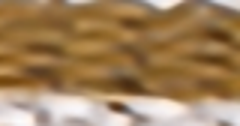}  &\hspace{-3.0mm}
			\includegraphics[width=0.15\linewidth, height = 0.165\linewidth]{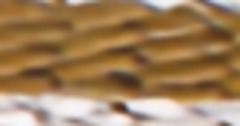} \\
			\multicolumn{3}{c}{~} &\hspace{-3.0mm}  (a) HR patch &\hspace{-3.0mm}  (b) Bicubic &\hspace{-3.0mm}  (c) VDSR~\cite{VDSR}  &\hspace{-3.0mm}  (d) ShuffleMixer~\cite{ShuffleMixer}\\
			
			\multicolumn{3}{c}{~} & \hspace{-3.0mm}
			\includegraphics[width=0.15\linewidth, height = 0.165\linewidth]{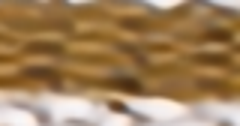} & \hspace{-3.0mm}
			\includegraphics[width=0.15\linewidth, height = 0.165\linewidth]{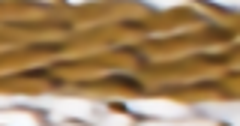} & \hspace{-3.0mm}
			\includegraphics[width=0.15\linewidth, height = 0.165\linewidth]{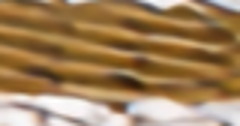} & \hspace{-3.0mm}
			\includegraphics[width=0.15\linewidth, height = 0.165\linewidth]{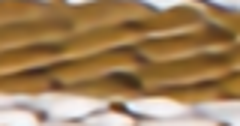} \\
			\multicolumn{3}{c}{\hspace{-3.0mm} img091 from Urban100} &\hspace{-3.0mm}  (e) LapSRN~\cite{LapSRN} &  \hspace{-3.0mm} (f) CARN~\cite{CARN}  &\hspace{-3.0mm}  (g) IMDN~\cite{IMDN}  & \hspace{-3.0mm}(h) SAFMN \\

            \\
            \multicolumn{3}{c}{\multirow{5}*[75pt]{\includegraphics[width=0.35\linewidth, height=0.355\linewidth]{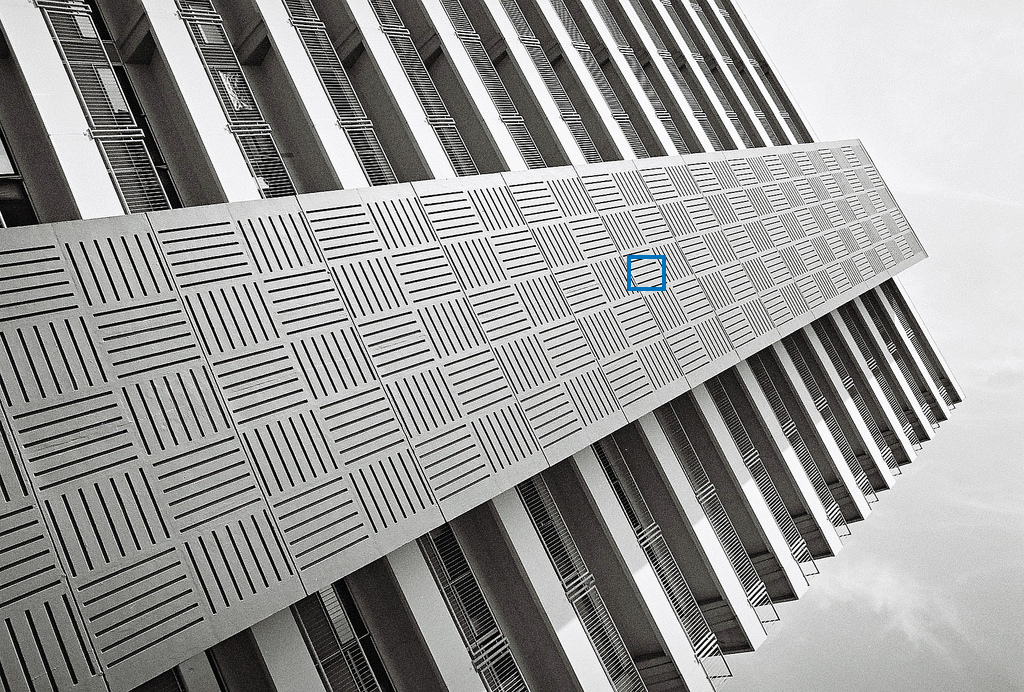}}}&\hspace{-3.0mm}
			\includegraphics[width=0.15\linewidth, height = 0.165\linewidth]{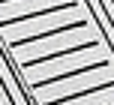} &\hspace{-3.0mm}
			\includegraphics[width=0.15\linewidth, height = 0.165\linewidth]{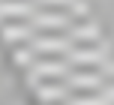}  &\hspace{-3.0mm}
			\includegraphics[width=0.15\linewidth, height = 0.165\linewidth]{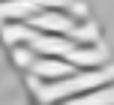}  &\hspace{-3.0mm}
			\includegraphics[width=0.15\linewidth, height = 0.165\linewidth]{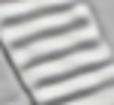} \\
			\multicolumn{3}{c}{~} &\hspace{-3.0mm}  (a) HR patch &\hspace{-3.0mm}  (b) Bicubic &\hspace{-3.0mm}  (c) VDSR~\cite{VDSR}  &\hspace{-3.0mm}  (d) ShuffleMixer~\cite{ShuffleMixer}\\
			
			\multicolumn{3}{c}{~} & \hspace{-3.0mm}
			\includegraphics[width=0.15\linewidth, height = 0.165\linewidth]{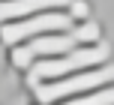} & \hspace{-3.0mm}
			\includegraphics[width=0.15\linewidth, height = 0.165\linewidth]{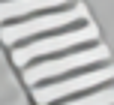} & \hspace{-3.0mm}
			\includegraphics[width=0.15\linewidth, height = 0.165\linewidth]{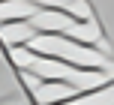} & \hspace{-3.0mm}
			\includegraphics[width=0.15\linewidth, height = 0.165\linewidth]{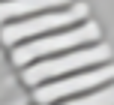} \\
			\multicolumn{3}{c}{\hspace{-3.0mm} img092 from Urban100} &\hspace{-3.0mm}  (e) LapSRN~\cite{LapSRN} &  \hspace{-3.0mm} (f) CARN~\cite{CARN}  &\hspace{-3.0mm}  (g) IMDN~\cite{IMDN}  & \hspace{-3.0mm}(h) SAFMN \\
			
		\end{tabular}
	\end{center}
	\vspace{-5mm}
	\caption{Visual comparisons for $\times4$ SR on the Urban100 dataset. Our method generates images with clearer structures.}
	\label{fig:visual_results}
	\vspace{-5mm}
\end{figure*}

In this section, we present additional visual comparisons with state-of-the-art methods~\cite{VDSR, CARN, LapSRN, IMDN, ShuffleMixer} on the $\times 4$ Urban100 dataset.
Figure~\ref{fig:visual_results} shows that the proposed algorithm generates clearer images with finer detailed structures than those by state-of-the-art methods.


{\small
	\bibliographystyle{ieee_fullname}
	\bibliography{egbib_sup}
}